\documentclass[11pt]{article}
\pdfoutput=1
\usepackage{amsmath}
\usepackage{amssymb}
\usepackage{amsthm}
\usepackage{amscd}
\usepackage{amsfonts}
\usepackage{graphicx}
\usepackage{fancyhdr}
\usepackage{color}
\usepackage{enumerate}
\usepackage{amsfonts}
\usepackage{amsthm}
\usepackage{psfrag} 
\usepackage{epsfig}
\usepackage{subfigure}
\usepackage{graphicx}
\usepackage{amssymb}
\usepackage{epstopdf}
\usepackage{amsbsy}
\usepackage{latexsym}
\usepackage{moreverb}  
\usepackage{booktabs}
\usepackage{textcomp}
\usepackage{bm}
\usepackage{soul}

\usepackage{multicol}
\usepackage{enumitem}

\usepackage{algorithm}
\usepackage{algorithmic}
\usepackage{hyperref}

\newtheorem{theorem}{Theorem}

\newtheorem*{example*}{Example}

\usepackage[top=2.8cm,bottom=2.8cm,left=2.5cm,right=2.5cm]{geometry}
\usepackage[T1]{fontenc}
\usepackage[utf8]{inputenc}

\usepackage[affil-it]{authblk} 
\usepackage{etoolbox}
\usepackage{lmodern}

\makeatletter
\patchcmd{\@maketitle}{\LARGE \@title}{\fontsize{18}{19.2}\selectfont\@title}{}{}
\makeatother

\usepackage{sectsty}
\sectionfont{\fontsize{13}{15}\selectfont}
\subsectionfont{\fontsize{12}{15}\selectfont}

\usepackage[textsize=tiny,textwidth=20mm]{todonotes}
\begin{document}

\title{Residual Multi-Fidelity Neural Network Computing}

\author[1]{Owen Davis\thanks{corresponding author: address:
University of New Mexico Department of Mathematics and Statistics
311 Terrace Street NE, Room 389 Albuquerque, NM, USA  87106,
 email: daviso@unm.edu}}
\author[1]{Mohammad Motamed\thanks{motamed@unm.edu}}
\author[2,3]{Ra{\'u}l Tempone\thanks{raul.tempone@kaust.edu.sa}}

\affil[1]{Department of Mathematics and Statistics, The University of
  New Mexico, Albuquerque, USA}
\affil[2]{Computer, Electrical and Mathematical Sciences and Engineering Division (CEMSE), King Abdullah University of Science and Technology (KAUST), Thuwal, Saudi Arabia}
\affil[3]{Alexander von Humboldt Professor in Mathematics for Uncertainty Quantification, RWTH Aachen University, Aachen, Germany
}

\date{}
\maketitle

\begin{center}
    {\it Dedicated to the memory of Ivo Bubu\v{s}ka}
\end{center}

\begin{abstract}

In this work, we consider the general problem of constructing a neural network surrogate model using multi-fidelity information. Motivated by error-complexity estimates for ReLU neural networks, we formulate the correlation between an inexpensive low-fidelity model and an expensive high-fidelity model as a possibly non-linear residual function. This function defines a mapping between 1) the shared input space of the models along with the low-fidelity model output, and 2) the discrepancy between the outputs of the two models. The computational framework proceeds by training two neural networks to work in concert. The first network learns the residual function on a small set of high- and low-fidelity data. Once trained, this network is used to generate additional synthetic high-fidelity data, which is used in the training of the second network. The trained second network then acts as our surrogate for the high-fidelity quantity of interest. We present four numerical examples to demonstrate the power of the proposed framework, showing that significant savings in computational cost may be achieved when the output predictions are desired to be accurate within small tolerances.
\end{abstract}

\medskip
\noindent
{\bf keywords:} 
multi-fidelity computing, residual modeling, deep neural networks, uncertainty quantification


\section{Introduction}
\label{sec:intro}

Deep artificial neural networks are widely regarded as central tools
to solve a variety of artificial intelligence problems;
see e.g. \cite{Schmidhuber:15} and the references therein. 
The application of neural networks is, however, not limited to artificial
intelligence. 
In recent years, deep networks have been successfully used to 
construct fast-to-evaluate surrogates for physical/biological quantities of
interest (QoIs) described by complex systems of ordinary/partial
differential equations (ODEs/PDEs); see
e.g. \cite{Sirignano_Spiliopoulos:18,E_Yu:18,E_etal:20}. This has resulted
in a growing body of research on the integration of physics-based modeling with
neural networks \cite{Willard_etal:20}. 
As surrogate models, deep networks have several important properties: 
1) their expressive power enables them to accurately approximate
 a wide class of functions; see e.g. \cite{Barron:93,Hornik_etal:89,Pinkus:99,Yarotsky:2017,Elbrachter_etal:2019,Daubechies_etal:2019,Guhring_etal:2020,Davis_Motamed:2024}; 
2) they can handle high-dimensional parametric problems involving many input
parameters; see e.g. \cite{Jentzen_etal:2018,Grohs_etal:2018,Kutyniok_etal:2019,Grohs_Herrmann:2020,Hutzenthaler_etal:2020}; and
3) they are fast-to-evaluate surrogates, as their evaluations mainly
require simple matrix-vector operations. 
Despite possessing such desirable properties, deep network surrogates are often very expensive to construct, especially when high levels of accuracy are desired \cite{adcock2021gap} and the underlying ODE/PDE systems are expensive to compute accurately. The construction of deep networks often requires abundant high-fidelity training data that in turn necessitates many expensive high-fidelity ODE/PDE computations.

To address this issue, we put forth a residual multi-fidelity computational framework. 
Given a low-fidelity and a high-fidelity computational model, we formulate the relation
between the two models in terms of a residual function. This function is defined as a possibly non-linear mapping between 1) the shared input space of the models together with the low-fidelity model output and 2) the discrepancy between the two model outputs. This formulation is motivated by a recent theoretical result \cite{Davis_Motamed:2024}, which shows that for a large class of bounded target functions with minimal regularity assumptions there exists a ReLU network whose approximation error is bounded above by a quantity proportional to the uniform norm of the target function and inversely proportional to the network complexity. As such, given a specified network error tolerance, and assuming the discrepancy between model outputs has a small uniform norm relative to the high-fidelity quantity, the residual framework allows the relationship between models of different fidelity to be learned within the desired accuracy using a network of lower complexity. 
This approach, in contrast to directly learning the mapping between models, reduces the amount of training data required, thus making it less costly to acquire. 
The trained network is then used as a surrogate to efficiently generate additional high-fidelity data. 
Finally, the enlarged set of originally available and newly generated high-fidelity data is used to train a deep network that learns, and acts as a fast-to-evaluate surrogate, for the high-fidelity QoI. We present four numerical examples to demonstrate the power of the proposed framework. In particular, we show that significant savings in computational cost may be achieved when the discrepancy between models is small in uniform norm and the output predictions are desired to be accurate within small tolerances. 

We remark as well that the developed framework is not limited to bi-fidelity modeling; it extends naturally to multi-fidelity modeling problems where the ensemble of lower-fidelity models are strictly hierarchical in terms of their predictive utility per cost. To accomplish this extension a sequence of networks learns a sequence of residuals that move up the model hierarchy.

\medskip

\noindent{\bf Related work.} 
This work introduces a neural network-driven computational framework that leverages multi-fidelity information and residual modeling, bolstered by recent bounds on network approximation error and complexity \cite{Davis_Motamed:2024}. 
Multi-fidelity modeling with neural networks is a quickly growing research area with numerous recent works; see e.g., \cite{Aydin_etal:19,Liu_Wang:19,MFNN_Karniadakis:20,MFNN_Motamed:20,meng2021multi,guo2022multi,song2022transfer,howard2023multifidelity,conti2023multi,partin2023multifidelity}. In \cite{Aydin_etal:19,Liu_Wang:19,MFNN_Motamed:20}, standard feedforward networks are used to directly learn the correlation between the low- and high-fidelity model, but make different a-priori assumptions concerning the form of this correlation. The authors in \cite{MFNN_Karniadakis:20,howard2023multifidelity} leverage physics-informed networks for this multi-fidelity learning task, and the latter further utilizes operator learning frameworks. In \cite{guo2022multi}, several neural network modeling frameworks are proposed that leverage connections between neural network approximation and Gaussian process regression. The authors in \cite{partin2023multifidelity} tackle this multi-fidelity learning problem using convolutional encoder/decoder networks, and in \cite{conti2023multi} multi-fidelity information is utilized together with long short-term memory networks. In \cite{meng2021multi} a multi-fidelity Bayesian neural network framework is developed, and finally, in \cite{song2022transfer} multi-fidelity information is harnessed via a transfer learning approach. Many of these modeling paradigms have been empirically successful, but theoretical analysis that clarifies under what modeling conditions they are performative is generally intractable. In this way, the present work is different. Our modeling paradigm is inspired by the ReLU network approximation error-complexity estimate proved in \cite{Davis_Motamed:2024}, which provides high-level guidelines concerning the modeling conditions required for the developed computational framework to be performative. Importantly, the purpose of this work is not to develop a neural network modeling paradigm and provide exhaustive numerical comparison with the existing methods in the literature. Rather, our main objective is to develop a theoretically supported modeling paradigm and verify numerically that it performs well under the predicted conditions.

Our work also leverages a type of residual modeling, but unlike most residual modeling approaches, where the residual is often defined as the difference of observed and simulated data (see e.g. \cite{Anscombe_Tukey:1963}), we define the residual as the discrepancy between high- and low-fidelity data. Our framework is also different from recent reduced-order modeling approaches (see e.g. \cite{San_Maulik:2018}) that utilize residuals to measure the discrepancy between full-order and reduced-order models. While in these approaches residuals account for model reduction due to modal
decomposition and truncation, i.e.~residuals gauge the discarded modes in the original full-order model, we define residuals as the difference between the target output QoIs at different levels of fidelity. A similar residual modeling strategy to our framework can
be found in residual networks (ResNets) \cite{ResNets:2016}, where a residual function is defined as the difference between the output and input of
a block of neural layers. Our framework adds a systematic inclusion
of multi-fidelity modeling to ResNets; see a more detailed discussion on ResNets in Section~\ref{sec:algorithm}.

The rest of the paper is organized as follows. In Section \ref{sec:problem-background}, we state the mathematical formulation of the problem. In Section~\ref{sec:rmfnn}, we present the residual multi-fidelity neural network algorithm and show how it is rigorously motivated by recently proved theoretical results in \cite{Davis_Motamed:2024}. In Section \ref{sec:theoretical_rationale}, we discuss sources of error and computational complexity in the developed algorithm. In Section \ref{sec:numerics}, we present four numerical examples that demonstrate the power of the proposed framework. Finally, in Section \ref{sec:conclusion}, we conclude and outline future works.

\section{Problem formulation and background}\label{sec:problem-background}

In this section, we formulate the problem and provide a brief overview of multi-fidelity modeling and ReLU neural networks, laying the necessary groundwork for the subsequent material.

\subsection{Problem statement}
\label{sec:problem}

Let $\bm{u}: \ (\bm{x}, \bm{\theta}) \in X \times \Theta \subset \mathbb{R}^{n} \times \mathbb{R}^d \mapsto \bm{u}(\bm{x};\bm{\theta}) \in \mathbb{R}^{m}$ be the solution to a (computationally involved) system of parametric ODEs/PDEs, parameterized by $\bm{\theta}\in \Theta\subset \mathbb{R}^d$. 
Further, let $Q: \Theta \rightarrow {\mathbb R}$ be a target function defined as a functional or operator, $M_Q$, acting on $\bm{u}$,
$$
Q:   \bm{\theta} \in \Theta \subset  {\mathbb R}^d  \mapsto
Q(\bm{\theta}) \in \mathbb{R}, \qquad Q({\bm{\theta}}) = M_Q(\bm{u}(\bm{x};\bm{\theta})).
$$ 
For example, the target function $Q$ may be given by the ODE/PDE solution energy, where $M_{Q}(\bm{u}(x;\bm{\theta})) = \int_{X}|\bm{u}(\bm{x};\bm{\theta})|^2d\bm{x}$. 
In general, since the target quantity $Q$ is available to us through a set of ODEs/PDEs, we may assume that it belongs to some Sobolev space.

Suppose that we need many accurate evaluations of $Q(\bm{\theta})$ at many distinct points $\bm{\theta} \in \Theta$. 
Such a situation arises, for example, in 
forward and inverse uncertainty quantification (UQ).
In principle, direct evaluations/realizations of $Q$ will require a very large number of expensive high-fidelity ODE/PDE computations that may be computationally prohibitive. Therefore, our goal is to construct a fast-to-evaluate surrogate of the target function $Q$, say $\tilde{Q}$, using only a small number of expensive
simulations of high-fidelity ODE/PDE models, and satisfying the accuracy constraint 
\begin{equation}\label{total_error}
\varepsilon := || Q - \tilde{Q} ||_{L_{\pi}^p(\Theta)} = \Bigl( \int_{\Theta} | Q(\boldsymbol\theta) - \tilde{Q}(\boldsymbol\theta) |^p \, d\pi(\boldsymbol\theta) \Bigr)^{1/p} \le \varepsilon_{TOL}.
\end{equation}

Here, we measure the approximation error in the $L^{p}$-norm, with $p \in [1, \infty)$ and weighted with a probability measure $d\pi$ on $\Theta$, and $\varepsilon_{TOL} \in (0,1/2)$ is a desired small tolerance. As we will show in Section \ref{sec:rmfnn}, we will achieve this goal by introducing a residual multi-fidelity framework that exploits the approximation power of deep networks and the efficiency of approximating quantities of small uniform norm. Once constructed, the surrogate can be combined with Monte Carlo (MC) and collocation methods (see e.g. \cite{Giles:11,CVMLMC_Nobile:15,MOMC:18,Multiindex:16}) to compute the statistics of the QoI; see the numerical examples in Section~\ref{sec:numerics}.

The new framework has two major components: i) multi-fidelity modeling; and ii) neural network approximation. We will briefly review these components in the remainder of this section.

\subsection{Multi-fidelity modeling}

The basic idea of multi-fidelity modeling is to leverage models at different levels of fidelity in order to achieve a desired accuracy with minimal computational cost. 
Let $Q_{LF}(\bm{\theta})$ and $Q_{HF}(\bm{\theta})$ be two approximations of the target quantity $Q(\bm{\theta})$, obtained by a low-fidelity and a
high-fidelity computational model, satisfying the
following two assumptions:
\medskip
\begin{itemize}
\item[{\bf A1}.] $Q_{HF}$ is a computationally expensive approximation of
$Q$, satisfying \eqref{total_error} with $\tilde{Q} =Q_{HF}$;

 \item[{\bf A2}.] $Q_{LF}$ is a computationally cheaper but less accurate approximation of $Q$, which is correlated with $Q_{HF}$.
\end{itemize}
\medskip
One common modeling scenario that can satisfy assumptions {\bf A1}-{\bf A2} occurs when the high- and low-fidelity models are built, respectively, from fine and coarse discretizations of the underlying ODE/PDE model. Specifically, suppose that
\begin{equation}\label{bi-fidelity}
Q_{LF}(\bm{\theta}) := M_{Q}(\bm{u}_{h_{LF}}(\bm{x}; \bm{\theta})), \qquad Q_{HF}(\bm{\theta}) = M_{Q}(\bm{u}_{h_{HF}}(\bm{x};\bm{\theta})), \qquad h_{HF} < h_{LF},
\end{equation}
where $h_{LF}$ and $h_{HF}$ denote the mesh size and/or the time step of a stable discretization scheme used at the low fidelity and high-fidelity levels, respectively. Assume further that 
\begin{equation}\label{HF_error}
| Q(\bm{\theta}) - Q_{LF}(\bm{\theta}) | \le c(\bm{\theta}) \, h_{LF}^q, \qquad
| Q(\bm{\theta}) - Q_{HF}(\bm{\theta}) | \le c(\bm{\theta}) \, h_{HF}^q, \qquad 
\forall \bm{\theta} \in \Theta,
\end{equation}
where $q>0$ is related to the order of accuracy of the discretization scheme, and $c = c(\bm{\theta})>0$ is a bounded function. Then this upper error bound
implies that assumptions {\bf A1}-{\bf A2} can be satisfied with proper choices of $h_{LF}$ and $h_{HF}$, e.g. with $h_{HF} \propto\varepsilon_{TOL}^{1/q}$ and $h_{LF}$ small enough for $Q_{LF}$ to be correlated with $Q_{HF}$. 

This example is intuitive and admits a very clear and mathematically rigorous notion of high- and low-fidelity, but importantly, it is not the only option. Lower-fidelity models can also be generated by solving an auxiliary problem obtained by simplifying the original problem. For instance, we may consider a model with simpler physics, an effective model derived through homogenization, or a reduced model obtained by smoothing out the rough parameters of $M_{Q}$, linearization, or modal decomposition and truncation. In this work, we primarily consider lower-fidelity models obtained by solving the underlying system of differential equations on a coarser discretization, but we additionally, in Section~\ref{sec:num_ex_1}, deploy the developed modeling framework on a test problem where the low-fidelity model is obtained by asymptotic approximation.

It is to be noted that in general the selection of low-fidelity and high-fidelity models is problem dependent. We consider a set of multi-fidelity models admissible as long as assumptions {\bf A1}-{\bf A2} are satisfied, but we note that satisfying {\bf A1}-{\bf A2} is not sufficient to guarantee that the chosen multi-fidelity models are optimal in terms of computational efficiency for a desired accuracy. 

The pivotal step in multi-fidelity computation is to design a procedure that captures and utilizes the relationship between models at different levels of fidelity. One of the most established approaches uses a Bayesian auto-regressive Gaussian process (GP) \cite{Kennedy_OHagan:2000, OND:GP2forrester2007}. This method works well for sparse high-fidelity data, but is often computationally prohibitive for high dimensional problems. It is in this high dimensional regime that neural network based multi-fidelity surrogate modeling promises to make an impact. A widely used neural network based approach, known as {\it comprehensive correction}, assumes a linear correlation between models, writing $Q_{HF}(\bm{\theta}) = \rho(\bm{\theta}) Q_{LF}(\bm{\theta}) + \delta(\bm{\theta})$, where $\rho$ and $\delta$ are the unknown multiplicative and additive corrections, respectively; see e.g. \cite{MF_review:16}. The main limitation of this strategy is its inability to capture a possibly nonlinear correlations between the two models. 
In certain regions of the domain $\Theta$, where the two models lack a strong linear correlation---such as in time-dependent problems where the low-fidelity data deteriorates and diverges from the high-fidelity data over time---a linear formulation may disregard the low-fidelity model (which could still provide valuable information) and assign all the weight to the high-fidelity data; see also Figure~\ref{pedagogy_fig}. In order to address this limitation, more general approaches have been proposed. In \cite{MF_nonlinear:17}, the authors replace the linear multiplicative relation by a nonlinear one and consider $Q_{HF}(\bm{\theta}) = F(Q_{LF}(\bm{\theta})) + \delta(\bm{\theta})$, where $F$ is a (possibly nonlinear) unknown function. Other related works that also employ neural networks to construct multi-fidelity regression models include
\cite{MFNN_Karniadakis:20,MFNN_Motamed:20}. In \cite{MFNN_Karniadakis:20}, a combination
of linear and nonlinear relations between models, $Q_{HF}(\bm{\theta}) = F_L (\bm{\theta}, Q_{LF}(\bm{\theta})) + F_{NL} (\bm{\theta}, Q_{LF}(\bm{\theta}))$ is used, and \cite{MFNN_Motamed:20} utilizes a general
correlation between the two models, $Q_{HF}(\bm{\theta}) = F(\bm{\theta}, Q_{LF}(\bm{\theta}))$, with $F$ being a general nonlinear unknown function. These formulations however are not capable of exploiting the full potential of multi-fidelity modeling to gain computational efficiency.

Motivated by residual modeling, instead of searching for a direct correlation between models, we reformulate the correlation in terms of a (possibly nonlinear) residual function that measures the discrepancy between the models; see Section \ref{sec:rmf}. We will show how and why such a formulation can achieve higher levels of efficiency.

\subsection{Neural network approximation}
\label{sec:relu_nn}

Deep neural network surrogates can be very expensive to construct \cite{adcock2021gap}, especially when high levels of accuracy are desired and the underlying ODE/PDE systems, that generate high-fidelity training data, are expensive to compute accurately. 
Although ResNets \cite{ResNets:2016} alleviate this issue to some extent by addressing the {\it degradation} problem (as depth increases, accuracy degrades), this work takes a step further by leveraging a combination of deep networks and cheap low-fidelity computations to efficiently build deep high-fidelity networks. 
We will specifically focus on ReLU feedforward networks, possibly augmented with a ResNet style architecture, as defined below.

\medskip
\noindent
{\bf ReLU networks.} Following the setup used in \cite{Petersen_Voigtlander:18,Guhring_etal:2020}, we define a feedforward network with $N_0=d$ inputs and $L$ layers, each with $N_{\ell}$ neurons, $\ell=1, \dotsc, L$, by a sequence of matrix-vector (or weight-bias) tuples
$$
\Phi := \{ (W_1,b_1), \dotsc, (W_L,b_L) \}, \qquad W_{\ell} \in
{\mathbb R}^{N_{\ell} \times N_{\ell - 1}}, \qquad
b_{\ell} \in {\mathbb R}^{N_{\ell}}, \qquad \ell=1, \dotsc, L.
$$
We denote by $f_{\Phi}$ the function that the network $\Phi$ realizes and write,
$$
f_{\Phi}(\boldsymbol\theta) = {\bf z}_L, \qquad f_{\Phi}: {\mathbb R}^d \rightarrow
{\mathbb R}^{N_L},
$$
where ${\bf z}_L$ results from the following scheme:
$$
{\bf z}_0 =\boldsymbol\theta, \qquad
{\bf z }_{\ell} = \sigma( W_{\ell} {\bf  z}_{\ell-1} + b_{\ell}), \ \ \ell=1, \dotsc, L-1, \qquad
{\bf z}_L  = W_{L} {\bf z}_{L-1} + b_{L}.
$$
Here, $\sigma: {\mathbb R} \rightarrow {\mathbb R}$ is the ReLU activation function $\sigma(z) = \max(0,z)$ that acts component-wise on all hidden layers, $\ell=1, \dotsc, L-1$. To make this ReLU network a ReLU ResNet we simply add shortcut connections to this structure which pass along the identity map.

\section{Residual multi-fidelity neural network algorithm}
\label{sec:rmfnn}

In this section, we present a residual multi-fidelity neural
network (RMFNN) algorithm consisting of two major parts: 1)
formulation of a new residual multifidelity framework, and 2)
construction of a pair of neural networks that work in concert within the new framework.
We will discuss each part in turn.

\subsection{Nonlinear residual multi-fidelity modeling}
\label{sec:rmf}

Combining residual analysis and multi-fidelity modeling, we
reformulate the mapping between two models of different fidelity as a non-linear residual function $F$ and write
\begin{equation}\label{corr_function}
Q_{HF}(\bm{\theta}) - Q_{LF}(\bm{\theta}) = F(\bm{\theta}, Q_{LF}(\bm{\theta})),
\qquad \bm{\theta} \in \Theta.
\end{equation}
As we will discuss below, this formulation enables the efficient
construction of a fast-to-evaluate high-fidelity surrogate by a pair of neural networks that work in concert. The first network, which learns the residual function $F$ is leveraged to train a second network that learns the high-fidelity target quantity $Q_{HF}$. The new formulation \eqref{corr_function} is motivated by the following two observations.

\begin{figure}[!htb]
\begin{center}
\includegraphics[width = 0.9\linewidth]{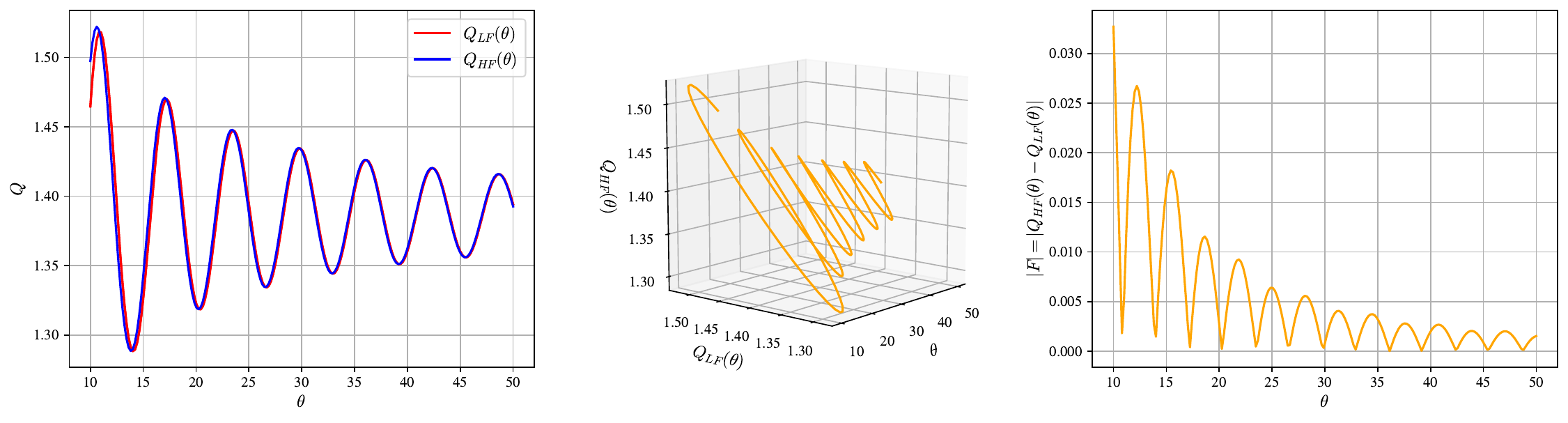}
\caption{Motivation behind the new residual multi-fidelity formulation. Left: deviation of the low-fidelity solution $Q_{LF}$ from the high-fidelity solution $Q_{HF}$, versus a frequency parameter $\theta \in \Theta=[10,50]$. Middle: high-fidelity solution is not a linear function of the low-fidelity solution on the whole parameter space. Right: residual function $F$ has a small magnitude.}
\label{pedagogy_fig}
\end{center}
\end{figure}

\medskip
\noindent
{\bf I. Nonlinearity.} \label{sec:non_linearity} 
In certain ODE/PDE problems, low-fidelity solutions lose their linear correlation with the high-fidelity solution on parts of the parameter domain. For example, in wave propagation problems, the low-fidelity solution may deteriorate as time increases due to dissipative and dispersive errors \cite{leveque2007finite}. The low-fidelity solution may also deteriorate at low frequencies if it is obtained by the truncation of an asymptotic expansion that is valid at high frequencies \cite{Condon_etal:2010}. 
As an illustrative example, consider a simple forced oscillator with damping given by the ODE system,
$$
\dot{\bf u}(t) = A \, {\bf u}(t) + \cos (\theta \,  t) \, {\bf b}, 
\qquad {\bf u} = \begin{bmatrix} 
	u_1 \\
	u_2\\
	\end{bmatrix},
\qquad
A= \begin{bmatrix} 
	0 & 1 \\
	-3 & -3\\
	\end{bmatrix},
\qquad {\bf b} = \begin{bmatrix} 
	0 \\
	0.6\\
	\end{bmatrix}.
$$
Here, $\theta \in \Theta=[10,50]$ is a frequency parameter. 
Let $Q(\theta) = u_2(T; \theta)^2$ be the desired quantity of interest, given
by the square of the second component of ODE's solution at a terminal time $T=1$. We use Forward Euler with a very small time step $(\Delta t = 10^{-6})$ to obtain a high-fidelity solution $Q_{HF}(\theta)$. 
For the low-fidelity model, we employ an asymptotic method (see \cite{Condon_etal:2010}), and noting
that $\cos (\theta \, t) = (e^{i \, \theta \, t} + e^{- i \, \theta \, t})/2$, we consider the ansatz
$$
{\bf u}(t) = {\bf u}_0(t) + \sum_{r = 1}^{\infty} \frac{1}{\theta^r} \,
{\bf u}_r(t; \theta), \qquad {\bf u}_r(t; \theta) = {\bf u}_{r,-1}(t) \, e^{-i \, \theta \, t} + {\bf u}_{r,1}(t) \, e^{i \, \theta \, t}. 
$$
Inserting this ansatz into the ODE system and matching the terms with the same $\theta^{-r}$ and $e^{\pm i \, \theta \, t}$ coefficients, we truncate the infinite sum and keep the first term with $r=1$. This yields the low-fidelity solution
$$
{\bf u}(t)={\bf u}_{LF}(t) + {\mathcal O}(\theta^{-2}), \qquad {\bf
  u}_{LF}(t) = e^{t A} \, {\bf u}(0) + \frac{1}{\theta} \, \sin (\theta \, t) \,
{\bf b}. 
$$
The high-fidelity solution is accurate but costly. The low-fidelity solution, having an explicit closed form, is cheap-to-compute, but it deteriorates when the frequency is low. Figure \ref{pedagogy_fig} (left) displays the deviation of $Q_{LF}$ from $Q_{HF}$ as the frequency decreases. In Figure \ref{pedagogy_fig} (middle) we observe that although the two
models exhibit a rather linear relation for larger frequencies, at lower frequencies ($\theta <30$) their relation is nonlinear. Hence for this system a purely linear auto-regressive approach would ignore $Q_{LF}$ (that is still informative) in a large portion of the parameter domain $\Theta$, instead putting all the weight on $Q_{HF}$. This would in turn increase the amount of (expensive) high-fidelity computations needed to build the surrogate model. On the contrary, the nonlinear formulation \eqref{corr_function} has the capability of capturing more complex mappings, fully exploiting $Q_{LF}$. 

\medskip
\noindent
{\bf II. Small residual magnitude.}\label{sec:small_res_mag}
It is important to note that we do not formulate the non-linearity discussed above as a direct relation between the two models, that is, we do not write $Q_{HF}(\boldsymbol\theta) = F(\boldsymbol\theta, Q_{LF}(\boldsymbol\theta))$. Instead, we formulate the non-linearity in terms of the residual \eqref{corr_function}. This is motivated by a recently proved approximation error estimate in \cite{Davis_Motamed:2024}, which we include here as Theorem~\ref{thm:main_approximation_error_result}. In order to state the theorem rigorously, we define the space of target functions 
\begin{equation}\label{eqn:approximation_space}
    S=\{Q:{\mathbb R}^d\to\mathbb{R}: ||Q||_{L^{1}(\mathbb{R}^d)}<\infty, \, ||\hat{Q}||_{L^{1}(\mathbb{R}^d)} < \infty, \, Q \not\equiv 0 \},
\end{equation}
where $\hat{Q}$ is the Fourier transform of $Q$. This space consists of all absolutely integrable functions on $\mathbb{R}^d$ with absolutely integrable Fourier transform, excluding the identically zero function. Importantly, functions in $S$ do not need to be continuous; they rather need to be continuous almost everywhere. We remark here that when $Q$ is discontinuous, we consider the relaxed notion of the Fourier transform \cite{grafakos2008classical}, where it is only assumed that
\begin{equation*}
    Q(\bm{\theta}) = \int_{\mathbb{R}^d}\hat{Q}(\bm{\omega})e^{i2\pi\bm{\omega} \cdot \bm{\theta}}\: d\bm{\omega}\qquad \text{ for almost every } \bm{\theta} \in \mathbb{R}^d.
\end{equation*}
\medskip
\begin{theorem}\label{thm:main_approximation_error_result}
Let $Q: \Theta \subseteq {\mathbb R}^d \rightarrow {\mathbb R}$ be a bounded function defined on the compact set $\Theta$, and assume $Q$ admits an extension $Q_{e}:\mathbb{R}^d\rightarrow\mathbb{R}$ such that $Q_{e}\in S$ and $||Q_{e}||_{L^{\infty}(\mathbb{R}^d)}\leq c||Q||_{L^{\infty}(\Theta)}$ for some $c\geq 0$.
For any $\varepsilon \in (0,1/2)$, there exists a ReLU network $Q_{\Phi}$ with depth $L\geq 2$, width $K > 2d+2$, and satisfying:
\begin{equation}\label{eqn:main_theoretical_tol}
    ||Q - Q_{\Phi}||_{L^{2}(\Theta)}\leq\varepsilon,
\end{equation}
such that 
\begin{equation}\label{eqn:main_theoretical_complexity_result}
    K \, L^{1/3} \leq C \, ||Q||^{2}_{L^{\infty}(\Theta)} \, \varepsilon^{-2} \, \log_2^{4/3}(\varepsilon^{-1}),
\end{equation}
where $C$ is a positive constant that may depend linearly on $d$ and logarithmically on $||Q||_{L^{\infty}(\Theta)}$ and $||\hat{Q}||_{L^{1}(\mathbb{R}^d)}$. 
\end{theorem}
\begin{proof}
    For the proof we refer to \cite{Davis_Motamed:2024}.
\end{proof}
Theorem~\ref{thm:main_approximation_error_result} guarantees the existence of ReLU networks approximating a large class of bounded target functions with minimal regularity assumptions to arbitrary accuracy. Moreover, the required complexity of the approximating networks can be bounded above by a quantity proportional to the uniform norm of the target function and inversely proportional to a quantity that exhibits polynomial dependence on the desired tolerance. We note as well that the constant depends logarithmically on $||\hat{Q}||_{L^{1}(\mathbb{R}^d)}$, which can be understood as a weak dependence on target function regularity. Precisely, the Fourier transform of a function $Q$ with integrable partial derivatives of order up to $m$ decays like $|\hat{Q}|\sim \mathcal{O}(|\bm{\omega}|^{-m})$ \cite{grafakos2008classical}, so the $L^{1}$ norm of the Fourier transform tends to increase with decreasing regularity. Importantly however, the constant $C$ scales logarithmically with respect to $||\hat{Q}||_{L^{1}(\Theta)}$, so as the uniform norm of the target function becomes small, the dependence of the network complexity on target function size is dominated by $||Q||^2_{L^{\infty}}$.

This estimate highlights the potential of the residual modeling framework, particularly when the discrepancy between low- and high-fidelity models is small in uniform norm relative to the high-fidelity model. Under such conditions, the residual function can be approximated by a lower-complexity network, which we hypothesize will require less expensive high-fidelity training data. 
We support this hypothesis numerically in Section~\ref{sec:numerics}. 
Notably, we do not claim that the discrepancy is always smaller in uniform norm but suggest that the framework is especially effective when this condition holds.

An example where the uniform norm of the residual function is small is the wave propagation problem discussed earlier; see Figure~\ref{pedagogy_fig}. 
As an additional example, consider a pair of low- and high-fidelity models obtained by coarse and fine discretizations of an ODE/PDE problem, satisfying
\eqref{HF_error}. Further, let $h_{LF} = s \, h_{HF}$ where $1<s < \infty$. Then using triangle inequality we can write
\begin{align*}
    | F | = |Q_{HF} - Q_{LF}| \le |Q - Q_{HF}| + |Q - Q_{LF}| &\le c (h_{HF}^q + h_{LF}^q )\\
    &\le c(1 + s^q) h_{HF}^q \le (1 + s^q) \, \varepsilon_{TOL}.
\end{align*}
The last inequality follows from the choice $ c \, h_{HF}^q \le\varepsilon_{TOL}$, which implies that $\varepsilon := | Q - Q_{HF}|\le \varepsilon_{TOL}$, as desired. Hence the size of the residual $|F|$ is proportional to the small quantity $\varepsilon_{TOL}$.

We remark here that Theorem~\ref{thm:main_approximation_error_result} only guarantees the existence of neural networks satisfying the  complexity estimate \eqref{eqn:main_theoretical_complexity_result}. It does not tell us how to find these networks. Hence an important assumption for the developed residual modeling framework is that we are able to successfully train neural networks that satisfy the approximation error estimate in Theorem~\ref{thm:main_approximation_error_result}.

In summary, the residual multi-fidelity formulation enables
the construction of an efficient pair of networks where the first network, which learns the residual function $F$, is leveraged to train a second network that learns the target quantity $Q_{HF}$; see Section \ref{sec:algorithm} for details. 

\subsection{Composite network training}
\label{sec:algorithm}

The second major component of the RMFNN algorithm involves constructing a pair of networks that work together. This construction proceeds in three steps.

\medskip
\noindent
{\bf Step 1. Data generation.} 
We generate a set of $N = N_I + N_{II}$ points, $\bm{\theta} \in \Theta$, collected in two disjoint sets: 
$$
\Theta_I := \{ \bm{\theta}^{(1)}, \dotsc, \bm{\theta}^{(N_{I})} \} \subset \Theta, \qquad \Theta_{II} := \{
  \bm{\theta}^{(N_{I}+1)}, \dotsc, \bm{\theta}^{(N)} \} \subset \Theta, \qquad \Theta_I
  \cap \Theta_{II} = \emptyset.
$$
The points in each set may be selected deterministically or randomly. 
For example, the two sets may consist of two uniform or non-uniform disjoint grids over $\Theta$, or we may consider a random selection of points drawn from a probability distribution, such as a uniform distribution or other optimally chosen distribution. 
We then proceed with the following computations.

\begin{itemize}
\item For each $\bm{\theta}^{(i)} \in \Theta_I \cup \Theta_{II}$, with $i=1,
  \dotsc, N$, we compute the low-fidelity realizations,
$$
Q_{LF}^{(i)} := 
Q_{LF}(\bm{\theta}^{(i)}) = M_{Q}(\bm{u}_{h_{LF}}(\bm{x};\bm{\theta}^{(i)})), \qquad i=1, \dotsc, N.
$$

\item For each $\bm{\theta}^{(i)} \in \Theta_{I}$, with $i=1,
  \dotsc, N_I$, we compute the high-fidelity realizations, 
\begin{equation}\label{QHF_realizations}
Q_{HF}^{(i)} := 
Q_{HF}(\bm{\theta}^{(i)}) = M_{Q}(\bm{u}_{h_{HF}}(\bm{x};\bm{\theta}^{(i)})), \qquad i=1, \dotsc, N_I.
\end{equation}
\end{itemize}

\begin{figure}[!h]
\begin{center}
\includegraphics[width = 0.7\linewidth]{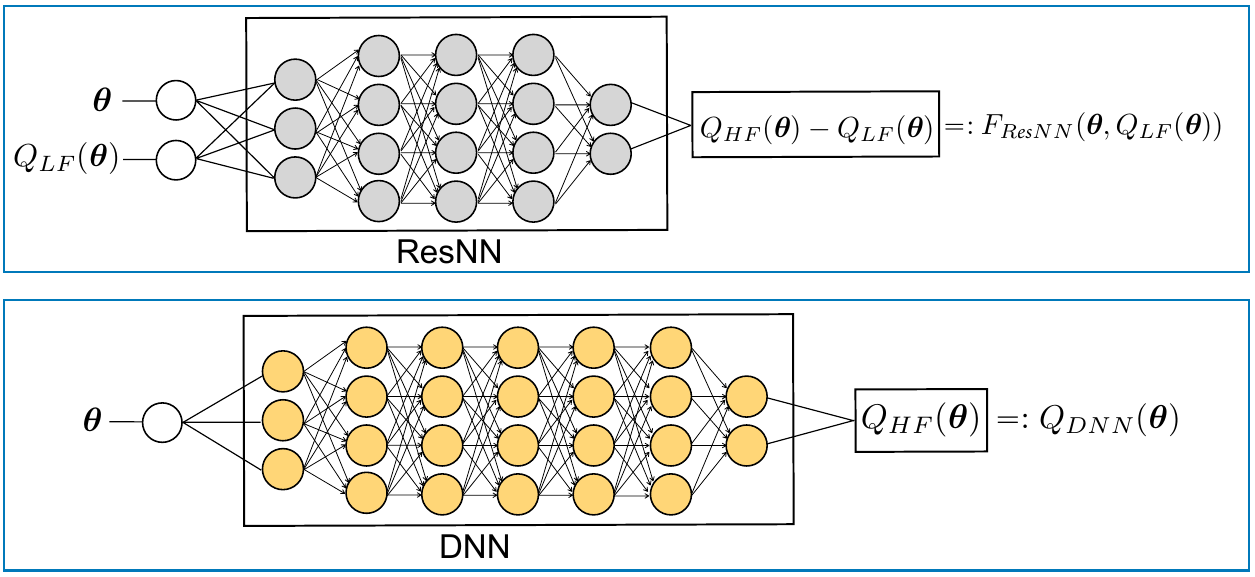} 
\caption{Schematic representation of the RMFNN algorithm, given
  a set of $N$ low-fidelity and $N_I \ll N$ high-fidelity data. Top: an initial network ($ResNN$) is trained by $N_I$ low-fidelity and
  high-fidelity data to learn the residual function $F$. The trained $ResNN$ is used along with the rest of $N_{II}=N-N_I$ low-fidelity data to generate a new set of $N_{II}$ high-fidelity data. Bottom: a deep network ($DNN$) is trained by all $N$ high-fidelity data as a surrogate for the high-fidelity
  target quantity $Q_{HF}$.}
\label{RMFNN_fig1}
\end{center}
\end{figure}

\medskip
\noindent
{\bf Step 2. Composite network training.} We train two separate networks as follows:
\begin{itemize}

\item Using $N_I$ input data, $\{ ( \bm{\theta}^{(i)},
  Q_{LF}^{(i)} )\}_{i=1}^{N_I}$ and $N_I$ output data, $\{Q_{HF}^{(i)} -
  Q_{LF}^{(i)} \}_{i=1}^{N_I}$, we train an initial network ($ResNN$) to learn the residual function $F$ in
  \eqref{corr_function}. We call this surrogate $F_{ResNN}$; see Figure \ref{RMFNN_fig1} (top).

\item We then use the surrogate $F_{ResNN}$ and the rest of the $N_{II}$ low-fidelity data points $\{ ( \bm{\theta}^{(i)}, Q_{LF}^{(i)} )\}_{i=N_I+1}^{N}$ to generate $N_{II}$ new approximate high-fidelity outputs, denoted by $\{ \bar{Q}_{HF}^{(i)} \}_{i=N_I+1}^{N}$, where 
\begin{equation}\label{QHF_new} 
\bar{Q}_{HF}^{(i)} := Q_{LF}^{(i)} + F_{ResNN}(\bm{\theta}^{(i)},
Q_{LF}^{(i)}), \qquad i=N_I+1, \dotsc, N.
\end{equation}

\item Using all $N$ input-output high-fidelity data $\{ (\bm{\theta}^{(i)}, Q_{HF}^{(i)}) \}_{i=1}^{N_I} \cup \{ ( \bm{\theta}^{(i)}, \bar{Q}_{HF}^{(i)})\}_{i=N_I+1}^{N}$, we train a deep network ($DNN$) as a surrogate for the high-fidelity quantity $Q_{HF}$. We call this surrogate $Q_{DNN}$; see Figure \ref{RMFNN_fig1} (bottom).

\end{itemize}

\noindent
{\bf Step 3. Prediction.} Given any $\bm{\theta} \in \Theta$, an approximation $\tilde{Q}(\bm{\theta})$ of the
target quantity $Q(\bm{\theta})$ can be obtained by evaluating the trained deep network $DNN$: 
\begin{equation}\label{predicted_QDNN}
\tilde{Q}(\bm{\theta}) = Q_{DNN}(\bm{\theta}), \qquad \bm{\theta} \in \Theta.
\end{equation}

\medskip
\noindent
{\bf On performance of RMFNN algorithm.} Theorem~\ref{thm:main_approximation_error_result} motivates the hypothesis that the RMFNN algorithm will be highly performative when $||F||_{L^{\infty}}\ll ||Q_{HF}||_{L^{\infty}}$. 
Given these conditions, it is important to note that the two networks in the algorithm learn two essentially different quantities: $ResNN$ learns the residual function $F$, while $DNN$ learns the target high-fidelity quantity ${Q}_{HF}$. 
Since $F$ has a smaller uniform norm compared to ${Q}_{HF}$, $ResNN$ can achieve the desired accuracy using a network with fewer degrees of freedom and less training data, relying on a smaller high-fidelity dataset of size $N_I$. In contrast, $DNN$ approximates $Q_{HF}$, which may require a deeper network and a larger dataset of size $N$. 
Importantly, this architectural difference between the two networks pairs well with the training data availability conditions assumed in this and other peer works on multifidelity modeling \cite{Aydin_etal:19,Liu_Wang:19,MFNN_Karniadakis:20,MFNN_Motamed:20,meng2021multi,guo2022multi,song2022transfer,howard2023multifidelity,conti2023multi,partin2023multifidelity}, where the number $N_I$ of available high-fidelity data is small compared to the number $N$ of available low-fidelity data, i.e. $N_{I}\ll N$. 
This distinction gives RMFNN an advantage over existing composite algorithms \cite{MFNN_Karniadakis:20,MFNN_Motamed:20}, where networks often learn quantities with similar uniform norms and require comparable high-fidelity training data sizes ($N_I \sim N$). By focusing on the residual, RMFNN reduces the reliance on large high-fidelity datasets, improving efficiency in multifidelity modeling.

\medskip
\noindent
{\bf An alternative approach.} We can take a slightly different approach as follows. The first step is the same as in Step 1 above, where we generate low-fidelity and high-fidelity data. 

\begin{figure}[!h]
\begin{center}
\includegraphics[width=0.7\linewidth]{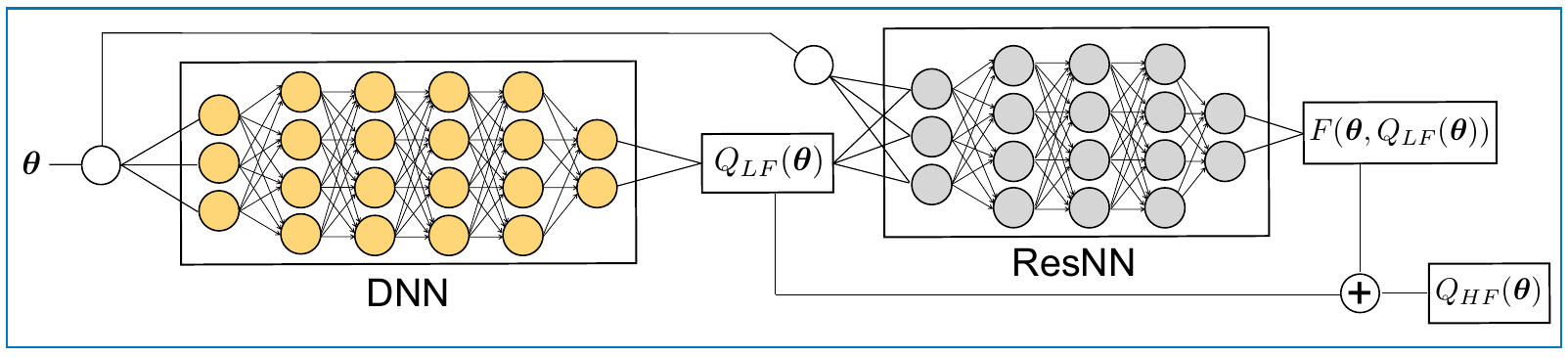}     
\caption{An alternative RMFNN approach. 
A deep network ($DNN$) is trained using $N$ low-fidelity data to learn $Q_{LF}$. 
A second network ($ResNN$) is trained by $N_I \ll N$ low-fidelity and high-fidelity data as a surrogate for the residual $F$. Finally, the target quantity $Q_{HF}$ is computed by adding $Q_{LF}$ to $F$.}
\label{RMFNN_fig2}
\end{center}
\end{figure}

Next, in Step 2, we construct a composite network as follows: 
\begin{itemize}
\setlength\itemsep{-.05em}
\item Using $N$ input-output data $\{ ( \bm{\theta}^{(i)}, Q_{LF}^{(i)}
  \}_{i=1}^{N}$, we train a deep network ($DNN$) as a surrogate for $Q_{LF}$.

\item Using $N_I$ input data $\{ ( \bm{\theta}^{(i)},
  Q_{LF}^{(i)} )\}_{i=1}^{N_I}$ and $N_I$ output data $\{Q_{HF}^{(i)} - Q_{LF}^{(i)} \}_{i=1}^{N_I}$, we train a second network ($ResNN$) as a surrogate for the residual function $F$ in \eqref{corr_function}. Note that this part is the same as the first part of step 2 in the original approach, but with a change in the order of implementation. 

\item We build a composite network by combining $DNN$ and $ResNN$ as shown in Figure \ref{RMFNN_fig2}. 
\end{itemize}

\noindent
For prediction in Step 3, given any $\bm{\theta}\in \Theta$, we compute an
approximation $\tilde{Q}(\bm{\theta})$ of the target quantity $Q(\bm{\theta})$ by adding the low-fidelity
quantity (predicted by $DNN$) to the residual (predicted by $ResNN$); see Figure \ref{RMFNN_fig2}.  

\medskip
\vspace{0.09cm}
\noindent
{\bf A comparison between the two proposed approaches.} Both approaches utilize the same residual multi-fidelity formulation \eqref{corr_function}, but they differ in the way their constituent networks ($ResNN$ and $DNN$) work. In the first approach, $ResNN$ and $DNN$ work at two separate stages: first, $ResNN$ learns the residual and generates more high-fidelity data; next, $DNN$ uses the initially available and newly generated high-fidelity data to learn the target quantity. In this case, $DNN$ alone serves as the surrogate network to be evaluated many times. In the alternative approach, $ResNN$ and $DNN$ constitute two blocks in a composite surrogate network, and hence both will be evaluated many times. In addition, since in both approaches $ResNN$ learns the same residual, and since $DNN$ learns either the low- or high-fidelity quantities, which are not expected to be very different in uniform norm, we expect the architectures of $ResNN$ and $DNN$ to be comparable across both approaches. This gives each approach distinct benefits. When we need many evaluations of the target quantity, the alternative approach may be more expensive, because it requires both $DNN$ and $ResNN$ to be evaluated, compared to only one evaluation of $DNN$ in the first approach. The alternative approach, however, has the flexibility of replacing $DNN$ by a direct computation of $Q_{LF}$, for instance when a direct computation of $Q_{LF}$ is more economical compared to the cost of training and evaluating a network.

\medskip
\noindent
{\bf Comparison with residual networks.} ResNets \cite{ResNets:2016} were originally developed to address the {\it degradation problem}, a phenomenon where we observe an eventual decrease in network training accuracy as depth increases. ResNets address this problem by introducing shortcut connections that pass along identity maps between layers. In this, as the network depth increases and accuracy becomes saturated, additional layers need only learn a residual that tends to zero as opposed to the full identity map plus some small possibly nonlinear correction. Understood in the context of the recent theoretical results in \cite{Davis_Motamed:2024}, ResNets take advantage of learning a residual that is small in uniform norm (relative to the identity map) between each layer. It is important to note that the size of the residual, as formulated in ResNet, is only guaranteed to be small relative to the identity map when the network accuracy is close to saturation and the identity map is near optimal. This situation arises when considering the impact of adding additional layers to a deep neural network that already approximates the quantity of interest to a reasonable accuracy (the conditions of the {\it degradation} problem). In the context of multi-fidelity modeling, where the map between low-fidelity and high-fidelity models needs to be learned accurately on sparse training data, it is unreasonable to expect the residual between layers (as formulated in ResNet) to be small compared to the identity map. Therefore, in order to take advantage of the previously mentioned theoretical results, we need to explicitly enforce learning a small quantity. This is the exact insight of our proposed RMFNN algorithm. We learn the map between models via a residual function $F$ that targets the discrepancy between high- and low-fidelity models, which in many cases is small in uniform norm.

\section{Error and complexity in RMFNN}\label{sec:theoretical_rationale}

In this section, we discuss sources of error in the RMFNN algorithm as well as its computational complexity.

\subsection{Sources of error} 

Let $\varepsilon$ denote the error \eqref{total_error} in approximating $Q$ by the surrogate \eqref{predicted_QDNN} obtained by the RMFNN algorithm, measured in $L^2$-norm, and satisfying the accuracy constraint
\begin{equation}\label{RMFNN_error}
\varepsilon = || Q - Q_{DNN} ||_{L^2(\Theta)} \le \varepsilon_{TOL}.
\end{equation}
By triangle inequality, we can split the error into three parts:
$$
\varepsilon \le \underbrace{|| Q - Q_{HF} ||_{L^2(\Theta)}}_{\varepsilon_I} 
+ \underbrace{|| Q_{HF} - \bar{Q}_{HF} ||_{L^2(\Theta)}}_{\varepsilon_{II}} 
+ \underbrace{|| \bar{Q}_{HF} - Q_{DNN} ||_{L^2(\Theta)}}_{\varepsilon_{III}} 
\le  \varepsilon_{TOL},
$$
where the desired accuracy is achieved if, for instance, we enforce $\varepsilon_I, \varepsilon_{II}, \varepsilon_{III} \le \frac{1}{3} \, \varepsilon_{TOL}$. The first error term $\varepsilon_I$ is the error in approximating $Q$ by $Q_{HF}$ in \eqref{QHF_realizations} using an ODE/PDE solver. Then, from \eqref{HF_error}, it follows that
$$
\varepsilon_I \le C \, h_{HF}^{q},
$$
ensuring the accuracy constraint is satisfied by choosing $h_{HF}$ such that $C \, h_{HF}^q \le \frac{1}{3} \, \varepsilon_{TOL}$. 
The second and third error terms are errors in network approximation, and hence of a different nature. Indeed, the second error term $\varepsilon_{II}$ is the error in approximating $Q_{HF}$ by $\bar{Q}_{HF}$ in \eqref{QHF_new}, which is the error in approximating $F$ in \eqref{corr_function} by the ResNN surrogate $F_{ResNN}$,
$$
\varepsilon_{II} = || Q_{LF} + F - Q_{LF} - F_{ResNN} ||_{L^2(\Theta)} = || F - F_{ResNN} ||_{L^2(\Theta)}.
$$
Similarly, the third error term $\varepsilon_{III}$ is the error in
approximating $\bar{Q}_{HF}$ by the DNN surrogate $Q_{DNN}$. Assuming infinite training data, Theorem~\ref{thm:main_approximation_error_result} guarantees the existence of neural networks $F_{ResNN}$, with depth $L_{1}$ and width $K_{1}$, and $Q_{DNN}$, with depth $L_{2}$ and width $K_{2}$, that satisfy the accuracy constraint and have complexity bounds given by 
\begin{align*}
    K_{1} \, L_{1}^{1/3} &\leq C_{1} \, ||F||^{2}_{L^{\infty}(\Theta)} \, \varepsilon_{TOL}^{-2} \, \log_2^{4/3}\left(\varepsilon_{TOL}^{-1}\right);\\
    K_{2} \, L_{2}^{1/3} &\leq C_{2} \, ||\bar{Q}_{HF}||^{2}_{L^{\infty}(\Theta)} \, \varepsilon_{TOL}^{-2} \, \log_2^{4/3}\left(\varepsilon_{TOL}^{-1}\right).
\end{align*}
Importantly, controlling $\varepsilon_{II}$ and  $\varepsilon_{III}$ also depends on our ability to identify the networks $F_{ResNN}$ and $Q_{DNN}$ through a training procedure using finite training data. To the best of the authors' knowledge, there are currently no neural network training paradigms that establish a rigorous relationship between the achieved tolerance and the quantity and distribution of training data. On the contrary,  neural networks are notoriously challenging to train, and the identification of networks that theoretical results, such as Theorem~\ref{thm:main_approximation_error_result}, guarantee to exist cannot be assumed; see, for example, \cite{adcock2021gap}. 
Thus, $\varepsilon_{II}$ and $\varepsilon_{III}$ should also be viewed as dependent on network training hyperparameters, as well as the quantity and distribution of training data, specifically the choices of the sets $\Theta_{I}$ and $\Theta_{II}$. 

Given that the training process is not rigorously understood, the ability to successfully train these networks must be verified numerically. In Section~\ref{sec:numerics}, we train all neural networks multiple times using different random seeds and report their variance. This approach aims to provide an empirical understanding of how uncertainty in neural network training manifests within the developed modeling paradigm.

Overall, when $F$ is small in uniform norm, Theorem~\ref{thm:main_approximation_error_result} guarantees the existence of a low-complexity network for which $\varepsilon_{II}$ is controlled. We pair this with the hypothesis that the number of training samples required to train a network decreases as its complexity decreases---a relationship we support numerically in Section ~\ref{sec:numerics}. Together, these motivate the developed modeling paradigm, suggesting that $F$ can be accurately approximated by a low-complexity network trained with a sparse set of both high- and low-fidelity data, which can then be employed to efficiently generate synthetic high-fidelity training data for the network $Q_{DNN}$.

%
%

\subsection{Computational complexity} 

In this section, we address the computational complexity of training and evaluating the networks $ResNN$ and $DNN$ under the accuracy constraint $\varepsilon_{II} + \varepsilon_{III} \le \frac{2}{3} \,  \varepsilon_{TOL}$.
We further emphasize that this work is specifically focused on leveraging neural networks as surrogates for expensive high-fidelity models within query heavy tasks such as uncertainty quantification, where the primary assumption is that evaluating the high-fidelity model represents the dominant cost that needs to be mitigated.

To this end, we discuss the computational complexity of the RMFNN algorithm, denoted as $W_{RMFNN}$, for approximating the target quantity $Q(\bm{\theta})$ at $N_{\bm{\theta}}$ distinct points, particularly when $N_{\bm{\theta}}\gg  1$ is large.

The complexity of the algorithm comprises two main components: the cost of training the two networks, $ResNN$ and $DNN$, and the cost associated with evaluations of the deep network estimator given by \eqref{predicted_QDNN}. The training costs for $ResNN$ and $DNN$ can be further broken down into the cost of generating training data and the cost of solving the optimization problem for the network parameters. Since we assume that the evaluations of the high-fidelity model are the primary cost to be reduced, we posit that the cost of training $ResNN$ and $DNN$ is dominated by the expense of obtaining high-fidelity training data. Thus, the dominant cost is assumed to be given by
\begin{equation}\label{RMFNN_cost}
W_{RMFNN} = N_I \, W_{HF} + N_{\bm{\theta}} \, W_{DNN},
\end{equation}
where $W_{HF}$ denotes the computational cost of one evaluation of $Q_{HF}$ in \eqref{bi-fidelity}, and $W_{DNN}$ denotes the cost of one evaluation of $DNN$. 
In what follows, we discuss the cost \eqref{RMFNN_cost} of RMFNN algorithm and compare it, for reference, with the costs associated with the following two approaches:

\begin{itemize}
\item[$\circ$] HFM: direct sampling of the high-fidelity model without utilizing neural
network-based surrogate construction or lower-fidelity information. Here the expensive
high-fidelity ODE/PDE model is directly computed $N_{\bm{\theta}}$ times with cost, 
\begin{equation}\label{HF_cost}
W_{HFM} = N_{\bm{\theta}} \, W_{HF}.
\end{equation}

\item[$\circ$] HFNN: evaluation of a deep high-fidelity neural network surrogate that is constructed without leveraging lower-fidelity information. The high-fidelity surrogate network is trained on only $N$ high-fidelity
data and evaluated $N_{\bm{\theta}}$ times with cost,
\begin{equation}\label{HFNN_cost}
W_{HFNN} = N \,  W_{HF} + N_{\bm{\theta}} \, W_{DNN}.
\end{equation}

\end{itemize}

The higher efficiency of the RMFNN method, compared to the above alternative approaches, relies on the following two conditions: 
\begin{center}
1) $N_I \ll N < N_{\boldsymbol\theta}$; \ \ \ \  and \ \ \ \ 2) $W_{DNN} \ll W_{HF}$. 
\end{center}
Clearly, these two conditions imply that $W_{RMFNN} \ll  W_{HFNN} < W_{HFM}$. 
We address each of these conditions in the two observations below.

\medskip
\noindent
{\it Observation 1}: 
When $||F||_{L^{\infty}}\ll ||Q_{HF}||_{L^{\infty}}$, Theorem~\ref{thm:main_approximation_error_result} guarantees the existence of networks $ResNN$, approximating $F$, and $DNN$, approximating $Q_{HF}$ where for a given tolerance, the complexity of $ResNN$ is considerably less than the complexity of $DNN$. Given this, we hypothesize that the number $N_I$ of expensive high-fidelity data needed to train $ResNN$ may be far smaller than the total number $N$ of data needed to train $DNN$, that is $N_I \ll N$. 
Moreover, in many scientific problems, we often need a very large number
($N_{\bm{\theta}} \gg 1$) of high-fidelity samples, implying $N < N_{\bm{\theta}}$. For example, when using Monte Carlo sampling to evaluate the statistical moments of random quantities, we require
$N_{\bm{\theta}} = {\mathcal O}(\varepsilon_{TOL}^{-2})$, since the statistical error in Monte Carlo sampling is known to be proportional
to $N_{\bm{\theta}}^{-1/2}$ \cite{MC}.
In such cases, the inequality $N < N_{\theta}$ holds provided $N = {\mathcal O}(\varepsilon_{TOL}^{-p})$, with $p<2$. We note that there are currently no rigorous results that relate the number and distribution of input training data to the accuracy of deep neural networks. However, we have observed through numerical experiments that for the type and size of networks that we have studied, the number of data needed to train a network to achieve accuracy within $\varepsilon_{TOL}$ indeed grows with a rate milder than ${\mathcal O} (\varepsilon_{TOL}^{-2})$.

\medskip
\noindent
{\it Observation 2}: The evaluation of a neural network mainly involves simple matrix-vector multiplications and evaluations of activation functions. The total evaluation cost depends on the width and depth of the network and the type of activation used. Crucially, the network evaluation cost depends only mildly on the complexity of the underlying ODE/PDE problem. As can be seen in Theorem~\ref{thm:main_approximation_error_result}, approximating target functions with successively lower regularity or larger problem input dimension will require neural networks of higher complexity (larger width and/or depth). However, when evaluating the neural network, this added complexity manifests only as additional simple matrix-vector operations and cheap evaluations of activation functions. 
In particular, for QoIs that can be well approximated by a network with several hidden layers and hundreds or thousands of neurons,
we expect $W_{ResNN} < W_{DNN} \ll W_{HF}$. Indeed, in such cases,
the more complex the high-fidelity problem, and the more expensive
computing the high-fidelity quantity, the smaller the ratio $W_{DNN}/W_{HF}$.

It is to be noted that in the above discussion we have assumed that
the cost of training $DNN$ is dominated by the cost of obtaining
high-fidelity training data, neglecting the one-time cost of solving
the underlying optimization problem in the training process. This will
be a valid assumption, for instance, when the cost of solving the
network optimization problem (e.g. by stochastic gradient descent) is
negligible compared to the cost of $N \gg1$
high-fidelity model evaluations. 
In particular, we have observed through numerical experiments that as the tolerance
decreases, and hence $N$ increases, this one-time network training cost may become negligible compared
to the cost of $N$ high-fidelity solves.  
We further illustrate this in
Section \ref{sec:numerics}. 

\medskip
\noindent
{\bf An illustrative example.} We compare the costs \eqref{RMFNN_cost},
\eqref{HF_cost}, and \eqref{HFNN_cost} on a prototypical UQ task. Let $\bm{\theta}$ be a
random vector with a known and compactly supported joint probability density function (PDF), and suppose that we want to obtain an accurate expectation
of the random quantity $Q=Q(\bm{\theta})$ within a small tolerance, $\varepsilon_{TOL} \ll 1$. To achieve this using MC sampling, recalling that the the statistical error in MC sampling is proportional to
$N_{\bm{\theta}}^{-1/2}$ \cite{MC}, we require 
$N_{\bm{\theta}} \propto \varepsilon_{TOL}^{-2} \gg 1$ samples. 
Now, suppose that $W_{HF} \propto h_{HF}^{- \gamma}$, where $\gamma>0$ is related
to the time-space dimension of the underlying ODE/PDE problem and the discretization
technique used to solve the problem. Then, noting $h_{HF} \propto
\varepsilon_{TOL}^{1/q}$, we obtain $W_{HF}
\propto \varepsilon_{TOL}^{-\gamma/q}$. 
Therefore, the cost \eqref{HF_cost} of directly computing the high-fidelity model
$N_{\bm{\theta}}$ times is
\begin{equation}\label{HFMC_cost_tol}
W_{HFM} \propto \varepsilon_{TOL}^{-(2+ \gamma/q)}.
\end{equation}
Following observation 1 above, we assume $N \propto
\varepsilon_{TOL}^{-p}$, with $p < 2$. This assumption indicates that the dependence of number $N$ of training data on the tolerance $\varepsilon_{TOL}$ is milder than the dependence of number of MC samples on the tolerance. Furthermore, we assume $r := N_I/N \ll 1$. Finally, following observation 2 above, we
assume that the evaluation cost of the deep network $DNN$ is less than the cost of a high-fidelity solve, i.e. $W_{DNN} \propto \varepsilon_{TOL}^{-\hat{p}}$, with $\hat{p} < \gamma/q$, implying $W_{DNN} \ll W_{HF}$. 
With these assumptions, the cost \eqref{HFNN_cost} of using a purely high-fidelity network surrogate and the cost \eqref{RMFNN_cost} of using a surrogate generated by the RMFNN algorithm read
\begin{equation}\label{HFNNMC_cost_tol}
W_{HFNN} \propto \varepsilon_{TOL}^{-
  (p+\gamma/q)} +\varepsilon_{TOL}^{-(2 + \hat{p})} + \text{training cost},
\end{equation} 
\begin{equation}\label{RMFNNMC_cost_tol}
W_{RMFNN} \propto r \, \varepsilon_{TOL}^{-
  (p+\gamma/q)} +\varepsilon_{TOL}^{-(2+ \hat{p})} +
\text{training cost}.
\end{equation} 
Comparing \eqref{HFMC_cost_tol} and \eqref{HFNNMC_cost_tol}, and excluding
training costs (see the discussion above), we get $W_{HFNN}  \ll
W_{HF}$, when $p<2$ and $\hat{p} < \gamma/q$. Furthermore, comparing \eqref{HFNNMC_cost_tol} and \eqref{RMFNNMC_cost_tol}, we get
$W_{RMFNN}  \ll W_{HFNN}$, when $r \ll 1$ and $\hat{p} < p+\gamma/q -2$.

\section{Numerical examples}
\label{sec:numerics}

In this section, we apply the proposed RMFNN algorithm to four numerical examples. The first one provides a focused numerical justification for the residual formulation present in the RMFNN algorithm, and the remaining examples comprise parametric ODE/PDE problems, of which the latter two are borrowed from \cite{MFNN_Motamed:20}. 
In all problems, the parameters are represented by a $d$-dimensional
random vector $\bm{\theta} \in \Theta \subset {\mathbb R}^{d}$, with a known and bounded joint PDF, $\pi:
\Theta \rightarrow {\mathbb R}_+$. 
For each parametric problem, we consider a desired output quantity $Q: \Theta
\rightarrow {\mathbb R}$, being a functional of the ODE/PDE solution. 
Our goal is to compute an accurate and efficient surrogate for $Q$,
denoted by $Q_{DNN}$, given by \eqref{predicted_QDNN}. 
We will measure the approximation error using the weighted $L^2$-norm, as defined in \eqref{total_error} with $p=2$. 
We approximate the error by sample averaging, 
using $N_{\varepsilon}$ independent samples $\{ \bm{\theta}^{(i)} \}_{i=1}^{N_{\varepsilon}} \in \Theta$ drawn from $\pi(\bm{\theta})$, 
and record the weighted mean squared error (MSE), 
\begin{equation}\label{MSE_error}
\varepsilon = \int_{\Theta} | Q(\bm{\theta}) - Q_{DNN}(\bm{\theta}) |^2 \,
\pi(\bm{\theta}) \, d\bm{\theta} \approx \frac{1}{N_{\varepsilon}}
\sum_{i=1}^{N_{\varepsilon}} | Q(\bm{\theta}^{(i)}) - Q_{DNN}(\bm{\theta}^{(i)}) |^2  =: \varepsilon_{MSE}.
\end{equation}

To facilitate a comparison of our algorithm with that of \cite{MFNN_Motamed:20}, two of our numerical examples involve computing the expectation of $Q$,
$$
{\mathbb E}[Q] := \int_{\Theta} Q(\bm{\theta})  \, \pi(\bm{\theta}) \, d\bm{\theta},
$$
by MC sampling. 
Due to the slow convergence of MC sampling, obtaining an accurate estimation may
require a very large number ($N_{\bm{\theta}} \gg 1$) of realizations of $Q$, each of which requires an expensive high-fidelity ODE/PDE solve. 
We will approximate all $N_{\bm{\theta}}$ realizations of $Q$ by the proposed
surrogate $Q_{DNN}$, built on far less than $N_{\bm{\theta}}$ number of ODE/PDE solves. 
This is an exemplary setting where we need many evaluations
of $Q_{DNN}$. 
To this end, we generate $N_{\bm{\theta}}$ independent
samples of $\bm{\theta}$, say $\{ \bm{\theta}^{(i)} \}_{i=1}^{N_{\bm{\theta}}}$, according to the joint PDF $\pi(\bm{\theta})$ and
approximate ${\mathbb E}[Q]$ by the sample mean of its approximated realizations computed by the RMFNN algorithm: 
\begin{equation}\label{A_RMFNNMC}
{\mathbb E}[Q] \approx {\mathcal A}_{RMFNN} := \frac1{N_{\bm{\theta}}}
  \sum_{i=1}^{N_{\bm{\theta}}} Q_{DNN} (\bm{\theta}^{(i)}).
\end{equation}
We compare the complexity of \eqref{A_RMFNNMC}
with that of a direct high-fidelity MC computation,
\begin{equation}\label{A_HFMC}
{\mathbb E}[Q] \approx {\mathcal A}_{HF} := \frac1{N_{\bm{\theta}}}
  \sum_{i=1}^{N_{\bm{\theta}}} Q_{HF} (\bm{\theta}^{(i)}).
\end{equation}
In addition to the mean squared error \eqref{MSE_error}, we also
report the absolute and relative errors in expectations,
\begin{equation}\label{abs_rel_error}
\varepsilon_{abs} := | {\mathbb E}[Q(\bm{\theta})] - {\mathcal
  A}|, \qquad
\varepsilon_{rel} := | {\mathbb E}[Q(\bm{\theta})] - {\mathcal
  A}|  / | {\mathbb E}[Q(\bm{\theta})] |, 
\end{equation}
where the estimator ${\mathcal A}$ is either ${\mathcal A}_{RMFNN}$ or ${\mathcal A}_{HF}$. 
We note that $\varepsilon_{abs}$ and $\varepsilon_{rel}$ contain both the deterministic error (due to approximating $Q$ by either $Q_{DNN}$ or $Q_{HF}$) and the statistical error (due to MC sampling of $Q_{DNN}$ or $Q_{HF}$). In the case of approximation $Q$ by $Q_{DNN}$, we can employ triangle inequality and write 
$$
    \varepsilon_{abs} \leq \bigl\lvert \mathbb{E}[Q(\bm{\theta}) - Q_{DNN}(\bm{\theta})] \bigr\rvert + \bigl\lvert \mathbb{E}[Q_{DNN}(\bm{\theta})]-\mathcal{A}_{RMFNN} \bigr\rvert,
$$    
where the first (deterministic) term is indeed the error measured in the weighted $L^1$-norm, as given in \eqref{total_error} with $p=1$. 

In all examples the closed-form solutions to the problems, and hence the true values of $Q(\bm{\theta})$, are known. We use these closed-form solutions to compute errors in the approximations, and we always compare the cost of different methods subject to the same accuracy constraint. 
All codes are written in Python and run on a single CPU.  
To construct neural networks we leverage Pytorch \cite{OND:Pytorch:https://doi.org/10.48550/arxiv.1912.01703} in examples \ref{sec:num_ex_0}  and \ref{sec:num_ex_1}, and Keras \cite{chollet2015keras} in examples \ref{sec:num_ex_2} and \ref{sec:num_ex_3}, both of which are open-source neural-network
libraries written in Python.
It is to be noted that all CPU times are  measured by {\tt time.process\_time()} in Python, taking average of tens of simulations. 

\subsection{A comparison of residual formulations}\label{sec:num_ex_0}

While formulation \eqref{corr_function} represents the residual as a function of both $\theta$ and $Q_{LF}$, it can alternatively be expressed solely in terms of $\theta$, isolating it from $Q_{LF}$. This leads to an alternative framework, referred to here as the Discrepancy Neural Network (DiscrepNN), which models the difference between the low- and high-fidelity outputs as a function of $\theta$. 
In this example, we compare the performance of these two frameworks, where:

\medskip
\begin{enumerate}
    \item RMFNN learns: $F(\theta, Q_{LF}(\theta)) = Q_{HF}(\theta)) - Q_{LF}(\theta),$
    \item DiscrepNN learns: $G(\theta)= Q_{HF}(\theta) - Q_{LF}(\theta).$ 
\end{enumerate}
\medskip

\begin{figure}[!htb]
    \centering
\includegraphics[width=0.9\linewidth]{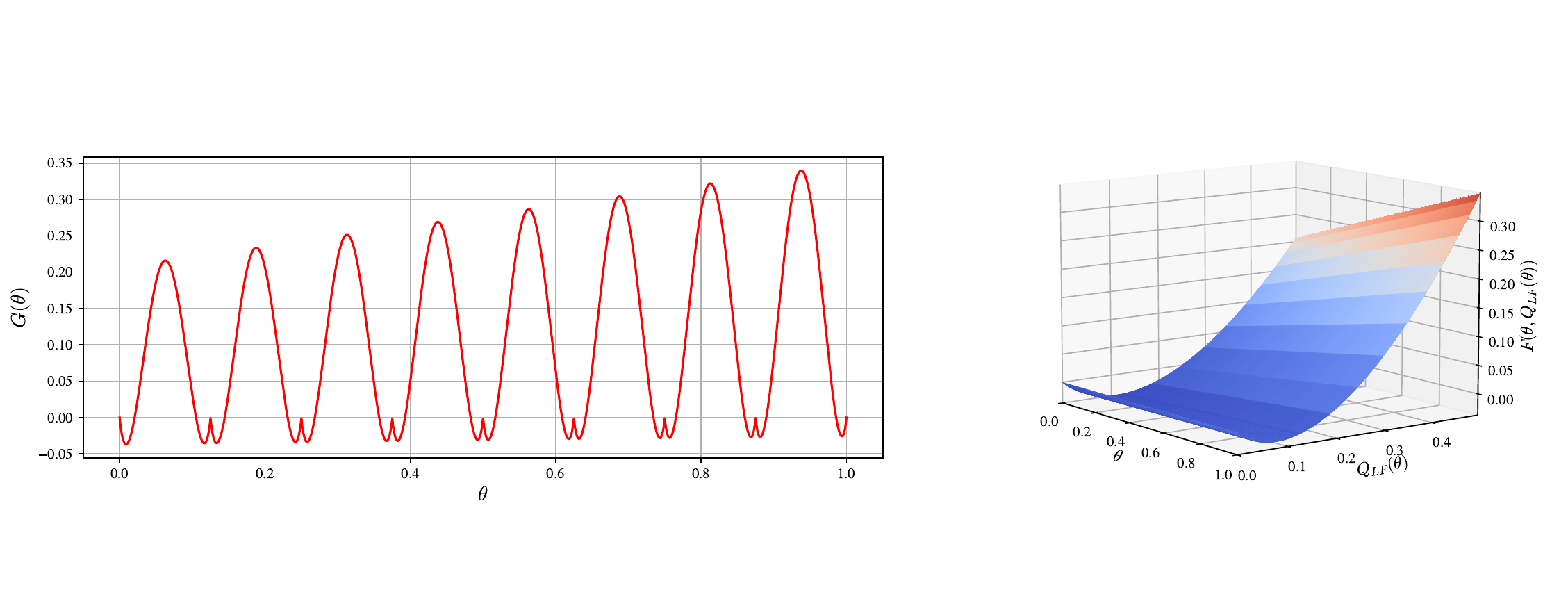}
    \caption{Two different residual functions: on the left $G(\theta)$, and on the right $F(\theta, Q_{LF}(\theta))$}
\label{fig:res_form_comp}
\end{figure}

Consider high- and low-fidelity models defined by
\begin{equation}
    Q_{LF}(\theta) = |0.5 \sin(8\pi(\theta+1))| \qquad Q_{HF}(\theta) =  2(1+0.2\theta)Q_{LF}(\theta)^{1.5}, \qquad \theta \in [0,1].
\end{equation}
In Figure~\ref{fig:res_form_comp}, we plot the residual function $G$ (left), and the residual function $F$ (right). 

\begin{figure}
    \centering
\includegraphics[width=0.45\linewidth]{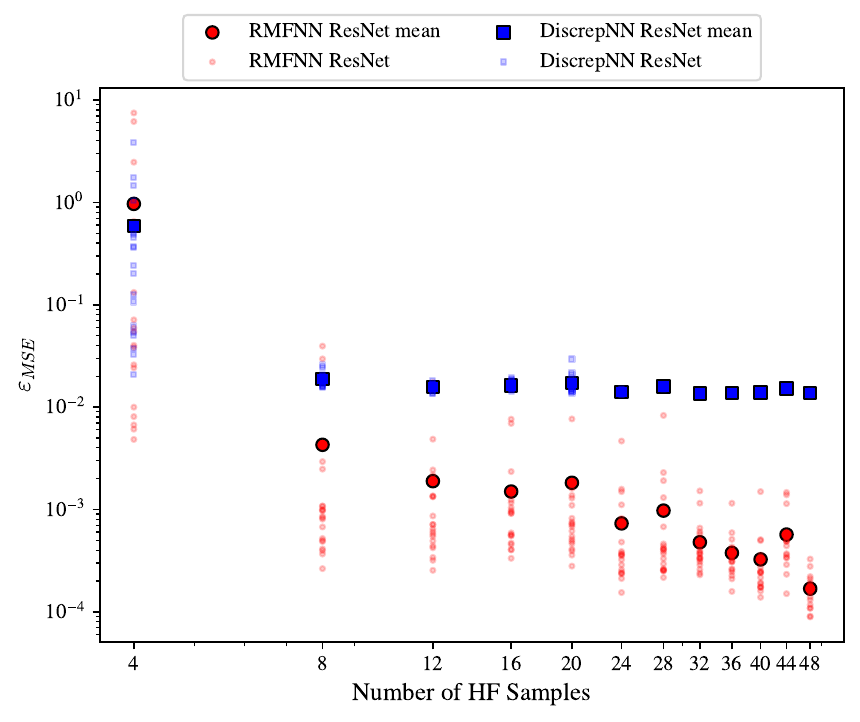}
    \caption{ Mean squared error $\varepsilon_{MSE}$ over 20 training runs approximating $Q_{HF}(\theta) - Q_{LF}(\theta)$ by RMFNN ResNet (red), and DiscrepNN ResNet (blue); average MSE over each ensemble is indicated by the larger marker with bold outline}
    \label{fig:RMFNN_vs_DiscrepNN}
\end{figure}

Note that while the input dimension of $F$ is one higher than that of $G$, $F$ is characterized by low frequency and a more regular profile. We hypothesize that this allows $F$ to be learned from sparser training data compared to $G$. To test this, we compare the approximation of 
$F$ and $G$ using neural networks under varying data sparsity conditions. 
Across both the RMFNN and DiscrepNN, we use a fixed neural network architecture consisting of $L=2$ layers and $K=10$ neurons per layer.
Moreover, all ReLU networks are constructed with a ResNet-style architecture, where we implement shortcut connections that pass along the identity map every two layers. We remark here that Theorem~\ref{thm:main_approximation_error_result} also holds for ReLU ResNets. This is shown in \cite{Davis_Motamed:2024}, where a strategy is developed to re-parameterize ReLU ResNets into standard ReLU networks with marginal change in network complexity.
For each method, the networks are trained on twelve different training sets where the number of high-fidelity (and low-fidelity) training samples ranges from $4$ to $48$. These training sets are generated using a random uniform distribution over the parameter space. Moreover, the training sample inputs and outputs are normalized so that they reside in $[0,1]\times[0,1]$. For training, we use the Adam optimization algorithm with the MSE cost function and Tikhonov regularization, and we implement an adaptive learning rate and stop criterion by monitoring the MSE on a validation set. To account for the randomness inherent in the training process, for each training set and for each surrogate method, we train and test 20 separate times on 20 distinct random seeds. From each train/test trial we compute the MSE \eqref{MSE_error} in the surrogate prediction on an independent test set of $N_{\varepsilon}=10^{6}$ points $\theta$ uniformly distributed in $[0,1]$. 
Figure~\ref{fig:RMFNN_vs_DiscrepNN} displays MSE versus the number of high-fidelity samples in the training set over 20 different training runs on distinct random seeds. The larger markers with bold outline indicate the average MSE over each ensemble.

In examining the sparsest training dataset (comprising only 4 samples), both the RMFNN and DiscrepNN exhibit similarly poor performance characterized by an average MSE around $10^{-1}$ and very high variance. This outcome is not surprising; with such limited data, it would be unexpected for either network to produce a high-quality surrogate. For the other datasets, the RMFNN consistently demonstrates a significantly lower average MSE compared to the DiscrepNN, suggesting that formulating the residual function as $F(\theta, Q_{LF}(\theta))$ is advantageous compared to the formulation $G(\theta)$. While the variance in the MSE for the RMFNN is generally higher than that of the DiscrepNN, it is noteworthy that this variability tends to favor predictions with a squared approximation error lower than the average across the ensemble. As the number of training samples increases, the average MSE for the DiscrepNN plateaus around $10^{-2}$, whereas the RMFNN's MSE continues to decrease, reaching nearly two orders of magnitude smaller at around $10^{-4}$. This performance discrepancy, coupled with the higher frequency content of the function $G$ compared to the function $F$, suggests that the DiscrepNN may require a more complex network architecture to further reduce MSE. This reinforces the importance of the residual formulation $F(\theta, Q_{LF}(\theta))$; within the RMFNN framework, the residual function can be effectively approximated by a low-complexity network capable of learning from sparse data. Overall, this example illustrates how incorporating low-fidelity model outputs into the residual function's input space can simplify the target function's profile, allowing it to be learned by a less complex network, even with limited training data.

\subsection{The impact of a small residual uniform norm}\label{sec:num_ex_1}

\begin{figure}[!htb]
    \centering
    \includegraphics[width = 0.4\textwidth]{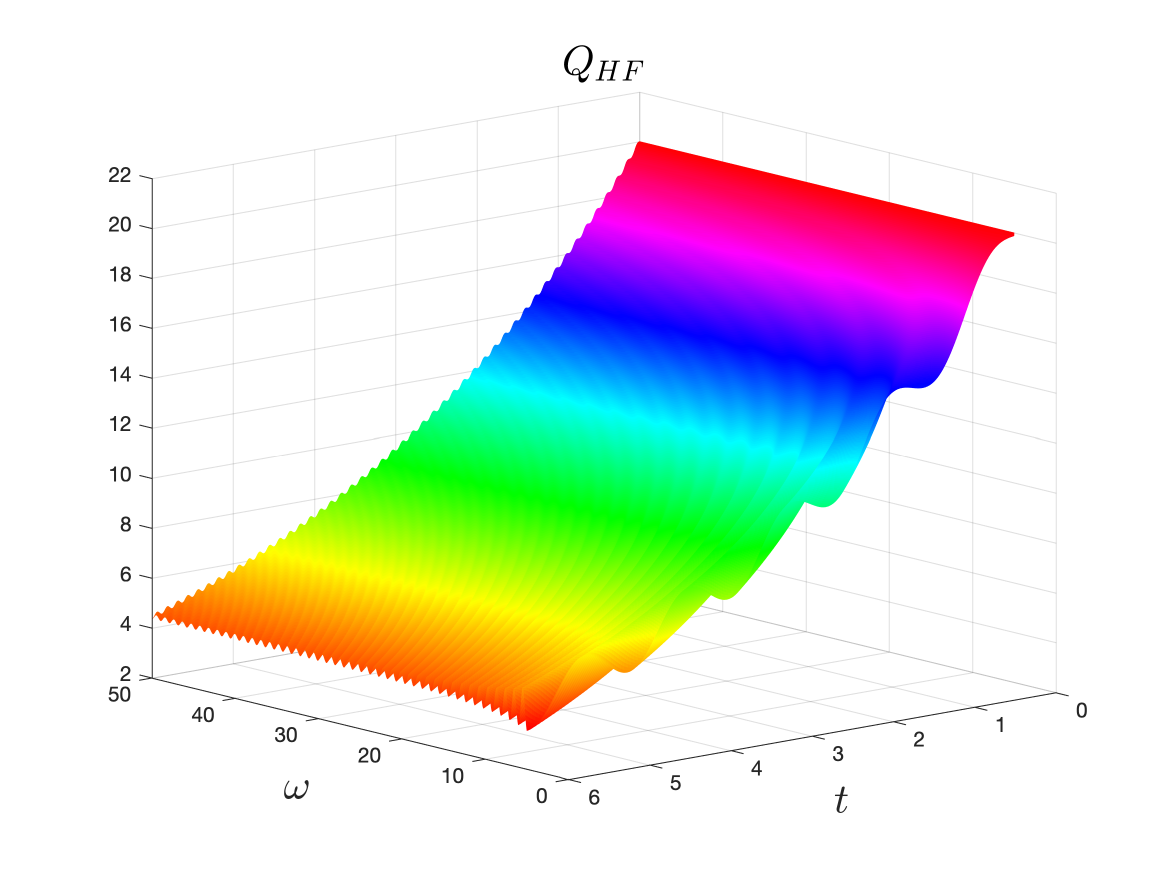}\includegraphics[width=0.4\textwidth]{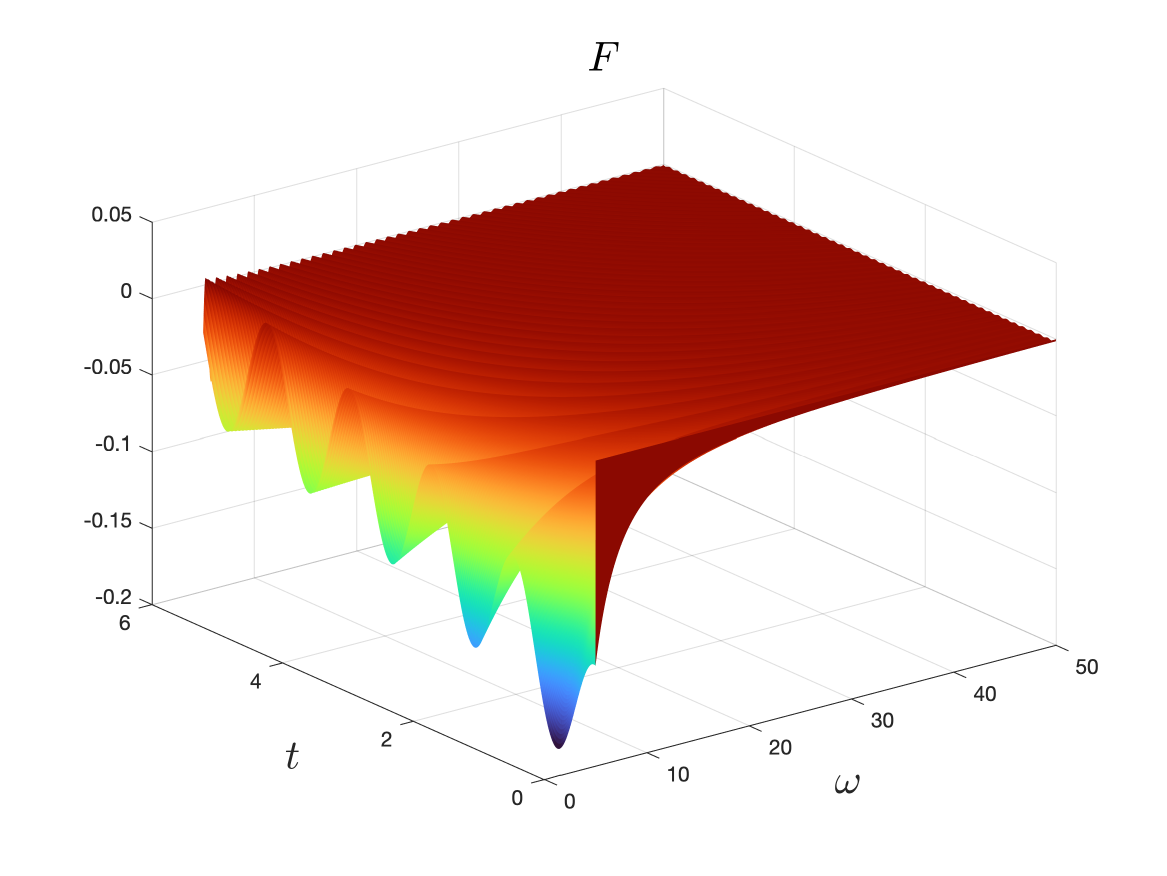}
    \caption{ Profiles of $Q_{HF}$ (left) and $F$ (right) versus $(\omega, t)$ for $b_1=0.2$ and $b_2=4.5$.}
    \label{fig:pulsed_oscillator_target_quantities}
\end{figure}

In this numerical experiment, we perform a surrogate construction task using bi-fidelity information, highlighting the advantage of approximating a residual function with small uniform norm.  
We consider a pulsed harmonic oscillator governed by the system of ODEs,
\begin{equation}\label{eqn:pulsed_oscillator}
  \dot{\bf u}(t) = A \, {\bf u}(t) + \cos (\omega \,  t) \, {\bf b}, 
\qquad {\bf u} = \begin{bmatrix} 
	u_1 \\
	u_2\\
	\end{bmatrix},
\qquad
A= \begin{bmatrix} 
	-2 & 0 \\
	  0 & -0.25\\
	\end{bmatrix},
\qquad {\bf b} = \begin{bmatrix} 
	b_1 \\
	b_2\\
	\end{bmatrix},
\end{equation} 
with initial solution $\mathbf{u}(0) = (1, 20)^{\top}$. We consider four model parameters: the frequency $\omega \in [5,50]$, the oscillation time $t \in [0,6]$, and two pulse parameters $(b_1, b_2) \in [0,0.2]\times[4,4.5]$, forming a four-dimensional parameter vector 
$\bm{\theta}= (\omega,t,b_1,b_2)$. 
Our goal is to construct a surrogate for the kinetic energy of the system, 
$$
Q(\bm{\theta})= || {\bf u}(\bm{\theta}) ||_2 = \bigl( u_{1}^{2}(\bm{\theta}) + u_{2}^{2}(\bm{\theta})\bigr)^{1/2}, \qquad \bm{\theta}= (\omega,t,b_1,b_2).
$$ 
To this end, we utilize multi-fidelity modeling as follows. 
For the high-fidelity model $Q_{HF}$ we leverage the exact analytic solution to \eqref{eqn:pulsed_oscillator}, and for the
low-fidelity model $Q_{LF}$ we employ an asymptotic method (see \cite{Condon_etal:2010} and section~\ref{sec:non_linearity}) to derive a closed form asymptotic approximation. In Figure~\ref{fig:pulsed_oscillator_target_quantities}, the residual $F=Q_{HF}-Q_{LF}$ and the high-fidelity quantity $Q_{HF}$ are plotted for $b_1 = 0.2$ and $b_2=4.5$. 
Notice that both $Q_{HF}$ and $F$ are of similar oscillatory character, but $||F||_{\infty}\ll ||Q_{HF}||_{\infty}$. Thus based on the discussion of Theorem~\ref{thm:main_approximation_error_result} in Section~\ref{sec:rmf}, we expect the RMFNN algorithm to be performative with the largest advantage under conditions of sparse training data. We will compare three different methods: 
\begin{enumerate}
    \item RMFNN ResNet: our proposed residual multi-fidelity algorithm using ReLU ResNets.
    \item MFNN ResNet: the multi-fidelity framework proposed in \cite{MFNN_Motamed:20} using ReLU ResNets; 
    \item HFNN ResNet: a single-fidelity ReLU ResNet trained on only high-fidelity data. 
\end{enumerate}

For the RMFNN algorithm we employ the alternative approach discussed in section~\ref{sec:algorithm}, and, further, we replace the training of the deep network $DNN$ by direct evaluations of the low-fidelity model. As such, each of the three surrogate methods requires the training of just one neural network. For clarity, we state the mapping that each of the methods learns below:
\begin{itemize}
    \item RMFNN learns: $(\bm{\theta},Q_{LF}(\bm{\theta})) \mapsto (Q_{HF}(\bm{\theta}) - Q_{LF}(\bm{\theta}))$
    \item MFNN learns: $(\bm{\theta}, Q_{LF}(\bm{\theta})) \mapsto Q_{HF}(\bm{\theta})$
    \item HFNN learns: $\bm{\theta}\mapsto Q_{HF}(\bm{\theta})$
\end{itemize}

\begin{figure}[!htb]
    \centering
    \includegraphics[width = 0.45\textwidth]{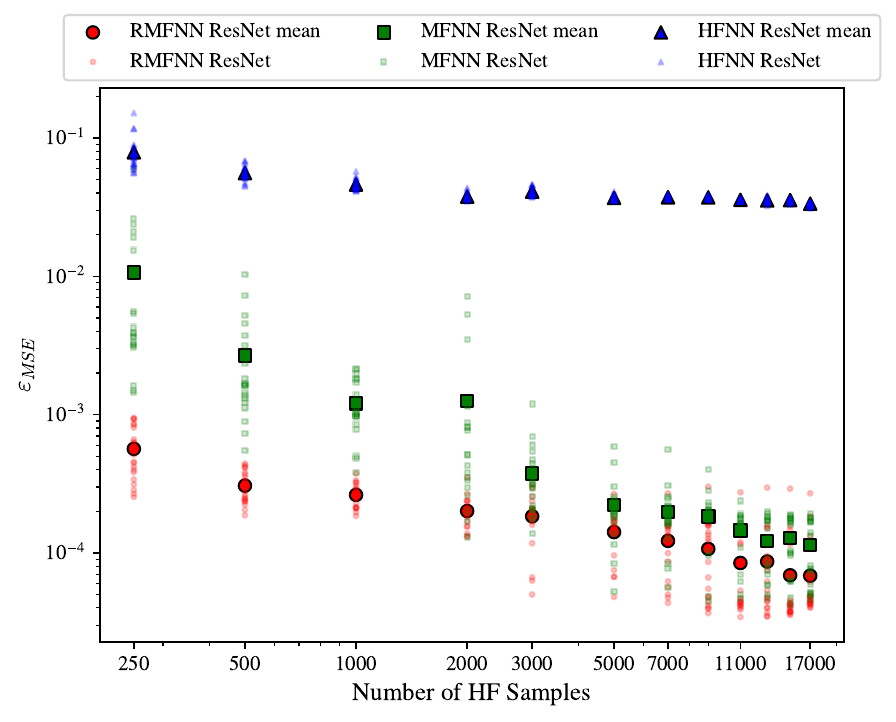}
    \caption{Mean squared error $\varepsilon_{MSE}$ over 20 training runs approximating $Q_{HF}$ by HFNN ResNet (blue), MFNN ResNet (green) and RMFNN ResNet (red) versus number of high-fidelity training samples; average MSE over each ensemble is indicated by the larger marker with bold outline}
    \label{fig:pulsed_oscialltor_results}
\end{figure}

Across all three methods we use a fixed ReLU network architecture consisting of $L=7$ layers, $K=7$ neurons in each hidden layer, and one ReLU-free neuron in the last layer. Moreover, to facilitate a comparison of our residual multi-fidelity framework with standard ResNets \cite{ResNets:2016}, all ReLU networks are constructed with a ResNet-style architecture, where we implement shortcut connections that pass along the identity map every two layers.
For each method, ReLU networks are trained on eleven different training sets where the number of high-fidelity (and low-fidelity) training samples ranges from $250$ to $17000$. These training sets are generated using a multi-variate uniform distribution over the parameter space. Moreover, the training sample inputs and outputs are normalized so that they reside in $[0,1]^{4}\times[0,1]$. For training, we use the Adam optimization algorithm with the MSE cost function and Tikhonov regularization. We implement an adaptive learning rate and stop criterion by monitoring the MSE on a validation set. To account for the randomness inherent in the training process, we train and test each surrogate method 20 separate times using distinct random seeds for each training set. 
From each train/test trial, we compute the MSE \eqref{MSE_error} in the surrogate prediction, evaluated on an independent test set of $N_{\varepsilon}=10^{6}$ points $\bm{\theta}$ uniformly distributed in $\Theta$. Figure~\ref{fig:pulsed_oscialltor_results} displays MSE versus the number of high-fidelity samples in the training set over 20 different training runs on distinct random seeds. The larger markers with bold outline indicate the average MSE over each ensemble.

As shown in Figure~\ref{fig:pulsed_oscialltor_results}, RMFNN consistently outperforms the other methods in terms of average MSE across the ensemble, with the performance advantage being most pronounced under sparse training data---the exact conditions desired in multi-fidelity computation. These results underscore the importance of incorporating both key principles that motivate the RMFNN algorithm:
\begin{enumerate}[label =(\roman*)]
    \item {\it Leveraging potentially non-linear correlations between low-fidelity and high-fidelity models.}
    \item {\it Ensuring the quantity learned during network training is small in uniform norm compared to the high-fidelity quantity of interest.}
\end{enumerate}
Table~\ref{tab:comp_RMFNN_MFNN_HFNN} summarizes the three surrogate methods in the context of properties (i) and (ii). A check mark (\checkmark) indicates that the method satisfies the corresponding property.

\begin{table*}[!htb]
    \centering
    \begin{tabular}{@{}lll@{}}
    \toprule
         methods & (i) & (ii)  \\
         \midrule
         HFNN &  & \\
         MFNN & $\checkmark$ & \\
         RMFNN & $\checkmark$ & $\checkmark$\\
         \bottomrule
    \end{tabular}
    \vspace{0.1cm}
     \caption{Comparison of RMFNN, MFNN, and HFNN methods in the context of properties (i) and (ii)}
    \label{tab:comp_RMFNN_MFNN_HFNN}
\end{table*}

\noindent Property (i) is harnessed by both RMFNN and MFNN, and the orders of magnitude reduction in mean squared error for the MFNN method over the HFNN method verifies its importance. 
Property (ii), on the other hand, is unique to RMFNN. 
The additional reduction in mean squared error achieved by RMFNN over MFNN validates the benefit of learning a quantity with a small uniform norm relative to the high-fidelity output.

The orders of magnitude reduction in error for the RMFNN ResNet over the HFNN ResNet also shows that our residual multi-fidelity framework makes a meaningful improvement over simply using ResNet-style shortcut connections in a single-fidelity network. In particular, these results support our discussion in Section~\ref{sec:algorithm}, where we compare the residual formulation in ResNet \cite{ResNets:2016} with our residual multi-fidelity framework.

 Examining the spread in MSE across different random seeds used for training, we observe significantly higher variance in the MFNN compared to the RMFNN in the sparse data regime. 
 For the HFNN and MFNN, the variance decreases as the number of high-fidelity training samples increases. In contrast, the variance in the RMFNN appears to increase slightly with larger training sets. Importantly, for more plentiful data, the variability in RMFNN predictions is biased towards MSE values smaller than the ensemble average, highlighting its robustness in performance.

We remark that while the results in Figure~\ref{fig:pulsed_oscialltor_results} were obtained for a fixed network architecture with $(K,L)=(7,7)$, the experiment was also conducted for $(K,L) = (25, 7)$ and $(K,L) = (7,15)$. The results obtained for these different network architectures are analogous to those pictured in Figure~\ref{fig:pulsed_oscialltor_results} up to a universal and commensurate reduction in MSE across all surrogate methods. Such a reduction in error can be explained by an increase in the number of trainable parameters and hence in the expressive capacity of the networks. 

\subsection{A parametric ODE problem}\label{sec:num_ex_2}

Consider the following parametric initial value problem (IVP),
\begin{equation}\label{IVP}
\begin{array}{ll}
u_t(t,\theta) + 0.5 \, u(t,\theta) = f(t,\theta), & \qquad t \in [0,T],  \\
u(0,\theta)=g(\theta),& 
\end{array}
\end{equation}
where $\theta \in \Theta= [-1,1]$ is a uniformly distributed random variable. Using the method of manufactured
solutions, we choose the force term $f$ and the initial data $g$ so that the exact solution to the IVP \eqref{IVP} is
$$
u(t,\theta) = 0.5+2 \sin(12\theta)+6 \sin(2t) \sin(10\theta) (1+2\theta^2).
$$

Now consider the target quantity 
$$
 Q(\theta) = u(T,\theta), \quad T=100.
$$
Our goal is to utilize the RMFNN algorithm to
construct a surrogate $Q_{DNN}$ for $Q$, which will then be used to perform MC sampling to estimate the expectation ${\mathbb E}[Q(\theta)]$. 

\medskip
\noindent 
{\bf Multi-fidelity models and accuracy constraint.} 
Suppose that we use the second-order accurate Runge-Kutta (RK2) time-stepper as the deterministic solver to compute realizations of $Q_{LF}(\theta)$ and $Q_{HF}(\theta)$ using time steps $h_{LF}$ and $h_{HF}$,
respectively. 
We perform a simple error analysis, similar to the analysis in
\cite{MFNN_Motamed:20}, to obtain the minimum number of realizations $N_{\theta}$ and the maximum time step $h_{HF}$ for the high-fidelity model
to satisfy the accuracy constraint $\varepsilon_{rel}  \le
\varepsilon_{TOL}$ with a $1\%$ failure
probability, i.e. ${\text{Prob}}(\varepsilon_{rel}  \le \varepsilon_{TOL}
) = 0.99$. Here $\varepsilon_{rel} $ is the relative error
given in \eqref{abs_rel_error}. Table \ref{Ex1_table1} summarizes the numerical parameters $(N_{\theta},
h_{HF}, h_{LF})$ and the CPU time of evaluating single realizations of
$Q_{LF}$ and $Q_{HF}$ for a decreasing sequence of tolerances
$\varepsilon_{TOL} = 10^{-2}, 10^{-3},
10^{-4}$. 
We choose $h_{LF} = s \, h_{HF}$, with $s=5,10,10$, for the three
tolerance levels.      
\begin{table*}[!h]
        \centering
        \caption{Required number of realizations and time steps to
        achieve ${\text{P}}( \varepsilon_{rel}  \le
\varepsilon_{TOL} ) = 0.99$.}
\medskip
        \label{Ex1_table1}
        \begin{tabular}{@{}llllll@{}}
\toprule
            $\varepsilon_{TOL}$ & $N_{\theta}$ & $h_{HF}$ &
          $W_{HF}$ & $h_{LF}$ & $W_{LF}$ \\
            \midrule
             $10^{-2}$ & $1.35 \times 10^{5}$ & 0.1     & $1.99 \times
                                                          10^{-4}$ & 0.5 & $3.99 \times
                                                          10^{-5}$\\
             $10^{-3}$ & $1.35 \times 10^{7}$ & 0.025 & $8.36 \times
                                                          10^{-4}$& 0.25 & $8.35 \times
                                                          10^{-5}$\\
             $10^{-4}$ & $1.35 \times 10^{9}$ & 0.01   & $2.39 \times
                                                          10^{-3}$& 0.1 & $2.39 \times
                                                          10^{-4}$\\
            \bottomrule
        \end{tabular}
\end{table*}

\noindent
{\bf Surrogate construction.} 
Following the algorithm in Section \ref{sec:algorithm}, we first
generate a set of $N= N_I + N_{II}$ points, $\theta^{(i)} \in [-1,1]$, with
$i=1, \dotsc, N$, collected into two disjoint sets, $\Theta_I$ and $\Theta_{II}$. 
We choose the points to be uniformly placed on the interval $[-1,1]$. We select every 10th point to be in the set $\Theta_I$, and we collect the rest of the points in the set $\Theta_{II}$. This implies $N \approx 10 \, N_I$, meaning that we need to compute the target quantity $Q(\theta)$ by the high-fidelity model
(using RK2 with time step $h_{HF}$) at only 10\% of points, i.e. $r=N_I/N
\approx 0.1$; compare this with $r=0.25$ taken in
\cite{MFNN_Motamed:20} to achieve the same accuracy. 
The number of points $N$ will be chosen based on the desired tolerance, slightly increasing as the tolerance decreases, i.e. $N \propto \varepsilon_{TOL}^{-p}$ where $p>0$ is
small. In particular, we observe in Table \ref{Ex1_table2} below that the value $p=0.5$ is enough to meet our accuracy constraints.

The architecture of $ResNN$
consists of 2 hidden layers, where each layer has 10 neurons. The architecture of $DNN$ consists of 4 hidden layers, where each layer has 20 neurons. The
architecture of both networks will be kept fixed at all tolerance levels. For the training process, we employ the MSE cost function and use Adam optimization technique.
For both networks, we split the available data points into a training
set (95$\%$ of data) and a validation set (5$\%$ of data) and adaptively tune the learning rate parameter. 
We do not use any regularization technique. Moreover, in order to take into account the uncertainty inherent in neural network training, each neural network is trained five separate times on five different random seeds.   
Table \ref{Ex1_table2}
summarizes the number of training-validation data $N_I$ (for $ResNN$) and $N$ (for $DNN$), the number of epochs $N_{epoch}$, batch size
$N_{batch}$, and the average CPU time of training and evaluating the two networks for different tolerances over the five independent training trials. We remark that the variance in training and evaluation time is negligible across the different trials. With these parameters we construct
three surrogates for $Q_{DNN}$, one at each tolerance level. 
We note that while using a different architecture and
other choices of network parameters (e.g. number of layers/neurons and
learning rates) may give more efficient networks, the selected
architectures and parameters here, following the general guidelines in
\cite{Bengio:12,GoodBengCour16}, produce satisfactory results in terms
of efficiency and accuracy; see Table \ref{Ex1_table_MSE} and Figures
\ref{Ex1_tol1_points}-\ref{Ex1_conv_tols} below. 
\begin{table*}[!h]
       \centering
        \caption{The number of training data and average training and
          evaluation time of the two networks.}
\medskip
        \label{Ex1_table2}
\resizebox{\textwidth}{!}{
\begin{tabular}{|c||c|c||c|c|c|c||c|c|c|c|}
\hline
& & &\multicolumn{4}{c||}{$ResNN$: 2$\times$10
      neurons}&\multicolumn{4}{c|}{$DNN$: 4$\times$20 neurons}\\
$\varepsilon_{TOL}$ & $N_I$ & $N$ & $N_{epoch}$ & $N_{batch}$ &  $W_{T_1}$ & $W_{P_1}$ &
                                                            $N_{epoch}$ & $N_{batch}$ & $W_{T_2}$ & $W_{P_2}$\\
\hline\hline
$10^{-2}$ & 25   & 241   &100& 10& 5.19& $2.10 \times
                                                          10^{-5}$&  400&40& 20.40& $2.80 \times 10^{-5}$\\
$10^{-3}$ & 81 & 801   &1500& 30& 70.08& $2.10 \times
                                                          10^{-5}$& 8000& 80& 446.39& $2.80 \times 10^{-5}$\\
$10^{-4}$ & 321 & 3201 &5000& 50& 276.86& $2.10 \times
                                                          10^{-5}$&20000& 200& 1378.82& $2.80 \times 10^{-5}$\\
\hline
\end{tabular}}
\end{table*}

Table \ref{Ex1_table_MSE} reports the MSE \eqref{MSE_error} in the constructed surrogates $Q_{DNN}$ using $N_{\varepsilon} = {10}^6$ evenly distributed
points $\theta$ in $\Theta$, confirming that the (deterministic) MSE in the network approximation is comparable to the (statistical) relative error $\varepsilon_{rel}$ over all five training runs.  
\begin{table*}[!h]
        \centering
        \caption{MSE in the approximation $Q_{DNN}(\theta) \approx Q(\theta)$.}
\medskip
        \label{Ex1_table_MSE}
        \begin{tabular}{llll}
\toprule
            $\varepsilon_{TOL}$ & $10^{-2}$& $10^{-3}$& $10^{-4}$\\
            \midrule
             $\varepsilon_{MSE}$  {\footnotesize(Trial 1)}& $6.98 \times 10^{-2}$ & $4.87\times 10^{-3}$ & $2.08\times 10^{-4}$ \\
             $\varepsilon_{MSE}$  {\footnotesize(Trial 2)}& $3.13 \times 10^{-2}$ & $5.71\times 10^{-3}$ & $1.37\times 10^{-4}$ \\
             $\varepsilon_{MSE}$  {\footnotesize(Trial 3)}& $5.10\times  10^{-2}$ & $7.66\times 10^{-3}$ & $2.95\times 10^{-4}$ \\
             $\varepsilon_{MSE}$  {\footnotesize(Trial 4)}& $9.10\times  10^{-2}$ & $5.50\times 10^{-3}$ & $1.97\times 10^{-4}$ \\
             $\varepsilon_{MSE}$  {\footnotesize(Trial 5)}& $1.14 \times 10^{-2}$ & $1.38\times 10^{-3}$ & $1.95\times 10^{-4}$ \\
             \bottomrule
        \end{tabular}
\end{table*}

Figure \ref{Ex1_tol1_points} shows the low-fidelity and high-fidelity
quantities versus $\theta \in [-1,1]$ (solid lines) and the data
(circle and triangle markers) available in the case $\varepsilon_{TOL}=10^{-2}$. 
Figure \ref{Ex1_toal1_predictions} (left) shows the pointwise mean of the generated
high-fidelity data by the network $ResNN$ over all five training runs together with 2 standard deviations. The variance in the generated data is very small, barely visible in the plot, owing to the fact that $ResNN$ predicts a quantity that is very small in uniform norm relative to the high-fidelity QoI. Figure
\ref{Ex1_toal1_predictions} (right) shows the pointwise mean predicted high-fidelity quantity by the deep network $DNN$ together with 2 standard deviations over the five independent training runs. 
\begin{figure}[!htb]
\begin{center}
\includegraphics[width=0.45\linewidth]{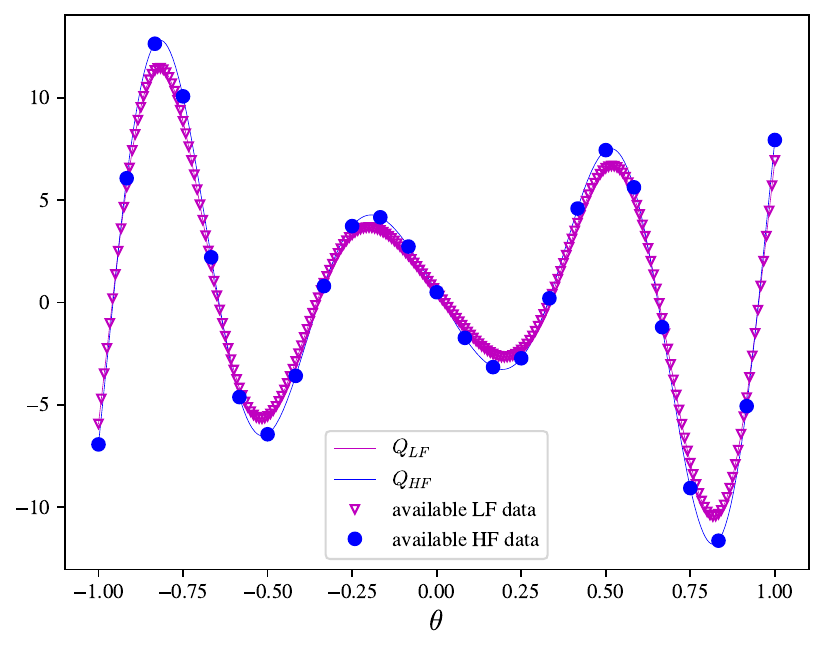}     
\caption{The low-fidelity and high-fidelity quantities versus
  $\theta\in[-1,1]$ (solid lines) and the
  available data (markers) in the case $\varepsilon_{TOL} =
  10^{-2}$. There are $N_I = 25$ high-fidelity and $N=241$
  low-fidelity data points, represented by circles and triangles, respectively.}
\label{Ex1_tol1_points}
\end{center}
\end{figure}
\begin{figure}[!htb]
\center
\includegraphics[width = 0.9\textwidth]{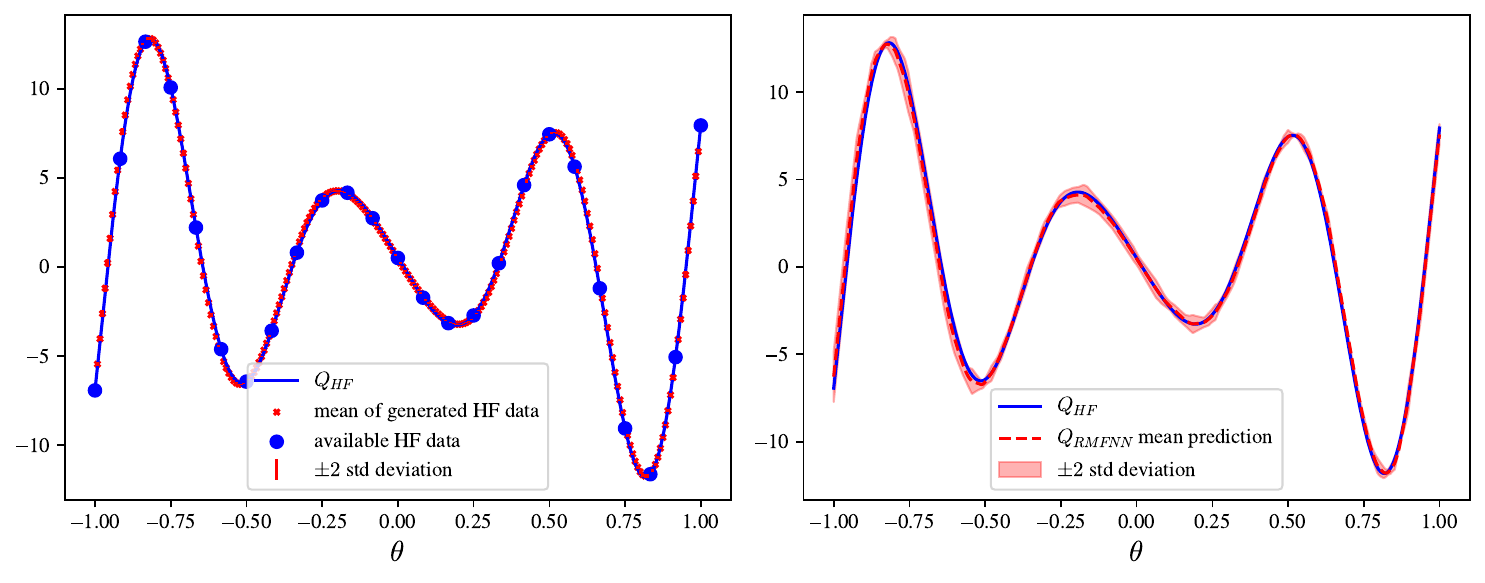}
    \caption{Outputs of the trained networks for $\varepsilon_{TOL} =
  10^{-2}$. Left: generated data by the network, $ResNN$. Right:
      predicted quantity by the deep network, $DNN$.}
    \label{Ex1_toal1_predictions}
 \end{figure}

\bigskip
\noindent
{\bf MC sampling.} We next compute ${\mathbb E}[Q]$ by
\eqref{A_RMFNNMC} and \eqref{A_HFMC} and compare their complexities. Figure \ref{Ex1_conv_tols} (left) shows the CPU time as a function of
tolerance. The computational cost of a direct high-fidelity MC
computation \eqref{A_HFMC} is ${\mathcal O}(\varepsilon_{TOL}^{-2.5})$, following \eqref{HFMC_cost_tol} and noting that the order of
accuracy of RK2 is $q=2$, and that the time-space dimension of the problem
is $\gamma = 1$.  On the other hand, if we only consider the prediction time of the
proposed residual multi-fidelity method, excluding the training costs,
the cost is ${\mathcal {O}(\varepsilon_{TOL}^{-2})}$, which is much less than the cost of high-fidelity MC sampling. When adding the training
costs, we observe that although for large tolerances the training cost
is large, as the tolerance decreases, the training costs become
negligible compared to the total CPU time. Overall, the cost of the
proposed method approaches ${\mathcal
  O}(\varepsilon_{TOL}^{-2})$ as tolerance
decreases, and hence, the smaller the tolerance, the more gain in
computational cost when employing the proposed method over
high-fidelity MC sampling. This can also be seen by
\eqref{RMFNNMC_cost_tol} noting that $\max(p + \gamma/q, 2+\hat{p}) =
\max(0.5 + 0.5,2+\hat{p}, ) = 2+ \hat{p}$, where $\hat{p}>0$ is very small. We note that while we observe significant cost reductions at the tested tolerances, training neural networks to achieve very small tolerances can be challenging, and what is required to do this in practice, be it additional training data or optimization time is not rigorously understood \cite{adcock2021gap}. Therefore, we do not assert that the demonstrated scaling will necessarily apply in these extremely small tolerance scenarios.
Finally, Figure \ref{Ex1_conv_tols} (right) shows 20 instances of the relative error as a function
of tolerance computed as 4 Monte Carlo simulations from each of the 5 independent network training runs. We see considerable variance in the network performance across different training runs, but even taking this variance into account, we meet the desired tolerance across all simulations.
\begin{figure}[!htb]
\centering
\includegraphics[width = 0.9\textwidth]{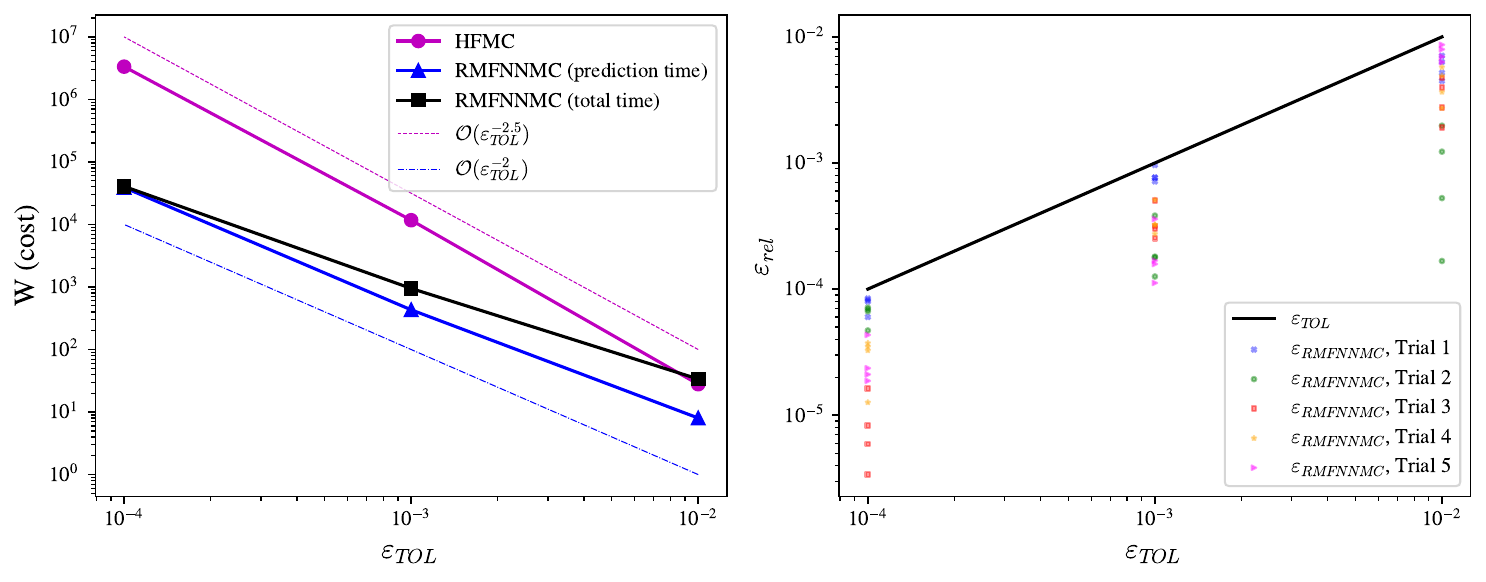}
\caption{CPU time and error versus tolerance. Left: for large tolerances the
  training cost is dominant, making the cost of
  RMFNNMC comparable to the cost of HFMC. However, as tolerance decreases,
  the training cost becomes negligible and the cost of
  RMFNNMC approaches ${\mathcal
  O}(\varepsilon_{TOL}^{-2})$; Right: relative error over different training trials as a function of tolerance verifies that the
  tolerance is met with 1$\%$ failure probability; different markers indicate 4 Monte Carlo simulations from each training trial}
\label{Ex1_conv_tols}
\end{figure}

\subsection{A parametric PDE problem}\label{sec:num_ex_3}

Consider the following parametric initial-boundary value problem
(IBVP)
\begin{equation}\label{IBVP}
\begin{array}{ll}
u_{tt}(t,\bm{x},\bm{\theta})- \Delta_{\bm{x}} u(t,\bm{x},\bm{\theta}) = f(t,\bm{x},\bm{\theta}),  & \ \ \
                                                                    (t, \bm{x},\bm{\theta}) \in [0,T] \times D \times \Theta,\\
u(0, \bm{x},{\bm{\theta}}) = g_1({\bm{x}},\bm{\theta}), \ \ u_t(0, \bm{x},\bm{\theta}) = g_2(\bm{x},\bm{\theta}), &  \ \ \ (t,\bm{x},\bm{\theta})
                                        \in \{ 0 \} \times D \times \Theta, \\
u(t, \bm{x},\bm{\theta}) = g_b(t,\bm{x},\bm{\theta}), &  \ \ \ (t,\bm{x},\bm{\theta})
                                        \in [0,T] \times \partial D \times \Theta,
\end{array}    
\end{equation}
where $t \in [0,T]$ is the time, $\bm{x} = (x_1, x_2) \in D$ is the vector of
spatial variables in a square domain $D=[-1,1]^2$, and $\bm{\theta} =
(\theta_1, \theta_2) \in \Theta$ is a vector of two independently and uniformly
distributed random variables on $\Theta = [10,11]\times[4,6]$. We
select the force term $f$ and the initial-boundary data $g_1, g_2, g_b$ so that the exact solution to the IBVP \eqref{IBVP} is
$$
u(t,\bm{x},\bm{\theta}) = \sin(\theta_1 \, t - \theta_2 \, x_1) \, \sin(\theta_2 \, x_2).
$$
We consider the target quantity, 
$$
Q(\bm{\theta}) = u(T,\bm{x}_Q, \bm{\theta}), \qquad T=30,
\qquad \bm{x}_Q = (0.5,0.5).
$$
Our goal is to construct a surrogate $Q_{DNN}$ for $Q$ and then to compute the expectation ${\mathbb E}[Q(\bm{\theta})]$ by MC sampling. 

\bigskip
\noindent
{\bf Multi-fidelity models and accuracy constraint.} 
Suppose that we have a second-order accurate (in both time and space) finite difference scheme as the deterministic solver to compute realizations of
$Q_{LF}(\bm{\theta})$ and $Q_{HF}(\bm{\theta})$ using a uniform grid with grid
lengths $h_{LF}$ and $h_{HF}$, respectively. 
We use the time step, $\Delta t = h/2$, to ensure
stability of the numerical scheme, where the grid length $h$ is either $h_{LF}$ or $h_{HF}$, depending on the level of fidelity. 
Given a $1\%$ failure probability and a decreasing sequence of tolerances $\varepsilon_{TOL} = 10^{-1}, 10^{-2}, 10^{-3}$,
a simple error analysis, similar to the analysis in
\cite{MFNN_Motamed:20}, and verified by numerical computations, gives
the minimum number of realizations $N_{\bm{\theta}}$ and the maximum grid length $h_{HF}$ for the high-fidelity model required to achieve $\text{Prob}(\varepsilon_{abs}  \le \varepsilon_{TOL}) = 0.99$. 
 Here, $\varepsilon_{abs} $ is the absolute error
given in \eqref{abs_rel_error}. 
Table \ref{Ex2_table1} summarizes the numerical parameters $(N_{\bm{\theta}},
h_{HF}, h_{LF})$, and the CPU time of evaluating single realizations of
$Q_{LF}$ and $Q_{HF}$. We choose $h_{LF} = s \, h_{HF}$, with $s=1.6,4,8$, for the three tolerance levels.  
\begin{table*}[!h]
        \centering
        \caption{Required number of realizations and grid lengths to
        achieve $\text{P}( \varepsilon_{abs}  \le
\varepsilon_{TOL} ) = 0.99$.}
\medskip
        \label{Ex2_table1}
        \begin{tabular}{@{}llllll@{}}
\toprule
            $\varepsilon_{TOL}$ & $N_{\bm{\theta}}$ & $h_{HF}$ &
          $W_{HF}$ & $h_{LF}$ & $W_{LF}$ \\
            \midrule
             $10^{-1}$ & $1.5 \times 10^{2}$ & $1/32$     & $0.29$ & $1/20$ & $0.075$\\
             $10^{-2}$ & $1.5 \times 10^{4}$ & $1/128$ & $10.50$& $1/32$ & $0.21$\\
             $10^{-3}$ & $1.5 \times 10^{6}$ & $1/320$   & $199.41$& $1/40$ & $0.38$\\
            \bottomrule
        \end{tabular}
\end{table*}

\bigskip
\noindent
{\bf Surrogate construction.} 
Following the RMFNN algorithm in Section \ref{sec:algorithm}, we first
generate a uniform grid of $N = N_I + N_{II}$ points $\bm{\theta}^{(i)} \in [10,11] \times [4,6]$, with
$i=1, \dotsc, N$, collected into two disjoint sets $\Theta_I$ and
$\Theta_{II}$. We select the two disjoint sets so that $N \approx 11 \, N_I$, meaning that we need to compute the quantity $Q(\bm{\theta})$ by the high-fidelity
model with grid length $h_{HF}$ at $9\%$ of points, i.e. 
$r=N_I/N \approx 0.09$; compare this with $r=0.25$ in
\cite{MFNN_Motamed:20}. 
The number $N$ of points will be chosen based on the desired
tolerance, slightly increasing as the tolerance decreases, i.e. $N
\propto \varepsilon_{TOL}^{-p}$ where $p>0$ is
small. In particular, as we observe in Table \ref{Ex2_table2} below,
the value $p=0.2$ is enough to meet our accuracy constraints.

The architecture of $ResNN$
consists of 2 hidden layers, where each layer has 20 neurons. The architecture of $DNN$
consists of 4 hidden layers, where each layer has 30 neurons. The
architecture of both networks will be kept fixed at all tolerance
levels. For the training process, we split the
available data points into a training set (95$\%$ of data) and a
validation set (5$\%$ of data). 
We apply pre-processing transformations to the input data points
before they are presented to the two networks. Precisely, we transform the
points from $[10,11] \times [4,6]$ into the unit square $[0,1]^2$. 
We employ the quadratic cost function and use the Adam optimization technique with an initial learning rate, $\eta = 0.005$. This learning rate is adapted during training by monitoring the MSE on the validation set. We do not use any regularization
technique.
Moreover, in order to take into account the uncertainty inherent in neural network training, each neural network is trained five separate times on five different random seeds.   
Table \ref{Ex2_table2}
summarizes the number of training-validation data $N_I$ (for $ResNN$) and $N$ (for $DNN$), the number of epochs $N_{epoch}$, batch size
$N_{batch}$, and the average CPU time of training and evaluating the two networks for different tolerances over the five independent training trials. We remark that the variance in training and evaluation time is negligible across the different trials. With these parameters we construct
three surrogates for $Q_{DNN}$, one at each tolerance level. 
\begin{table*}[!h]
       \centering
        \caption{The number of training data and training and
          evaluation time of the two networks}
\medskip
        \label{Ex2_table2}
\resizebox{\textwidth}{!}{
\begin{tabular}{|c||c|c||c|c|c|c||c|c|c|c|}
\hline
& & &\multicolumn{4}{c||}{$ResNN$: 2$\times$20 neurons}&\multicolumn{4}{c|}{$DNN$: 4$\times$30 neurons}\\
$\varepsilon_{TOL}$ & $N_I$ & $N$ & $N_{epoch}$ & $N_{batch}$ & $W_{T_1}$ & $W_{P_1}$ 
                                                                               & $N_{epoch}$ & $N_{batch}$ & $W_{T_2}$ & $W_{P_1}$\\
\hline\hline
$10^{-1}$ & 324   & 3498   &100& 50& 6.79& $2.08 \times10^{-5}$& 500&50&62.84& $5.80 \times 10^{-5}$   \\
$10^{-2}$ & 451 & 4961   &200& 50& 11.50& $2.08 \times10^{-5}$ 
                                          & 1000& 100& 105.04& $5.80 \times 10^{-5}$   \\
$10^{-3}$ & 714 & 8003 &1000& 50 & 58.00& $2.08 \times10^{-5}$ 
                                         &5000& 200& 483.46& $5.80 \times 10^{-5}$  \\
\hline
\end{tabular}}
\end{table*}

Table \ref{Ex2_table_MSE} reports the MSE \eqref{MSE_error} in the constructed surrogates $Q_{DNN}$, using a
uniform grid of $N_{\varepsilon} = {10}^6$ points $\bm{\theta}$ in $\Theta$, confirming that the (deterministic) MSE in the network approximation is comparable to the (statistical) absolute error $\varepsilon_{abs}$ over all five training runs.   

\begin{table*}[!h]
        \centering
        \caption{MSE in the approximation $Q_{DNN}(\theta) \approx Q(\theta)$.}
\medskip
        \label{Ex2_table_MSE}
        \begin{tabular}{llll}
\toprule
            $\varepsilon_{TOL}$ & $10^{-1}$& $10^{-2}$& $10^{-3}$\\
            \midrule
             $\varepsilon_{MSE}$  {\footnotesize(Trial 1)}& $2.05\times 10^{-3}$& $8.47\times 10^{-4}$ & $4.20\times 10^{-4}$ \\
             $\varepsilon_{MSE}$  {\footnotesize(Trial 2)}& $1.38 \times 10^{-3}$ & $2.04\times 10^{-3}$ & $1.22\times 10^{-4}$ \\
             $\varepsilon_{MSE}$  {\footnotesize(Trial 3)}& $4.64\times  10^{-3}$& $2.84\times 10^{-3}$ & $4.34\times 10^{-4}$ \\
             $\varepsilon_{MSE}$  {\footnotesize(Trial 4)}& $2.56\times  10^{-3}$& $3.21\times 10^{-3}$& $5.45\times 10^{-4}$ \\
             $\varepsilon_{MSE}$  {\footnotesize(Trial 5)}& $2.19 \times 10^{-3}$& $1.08\times 10^{-3}$& $1.03\times 10^{-3}$ \\
             \bottomrule
        \end{tabular}
\end{table*}

\begin{figure}[!htb]
\centering
\includegraphics[width =0.9\linewidth]{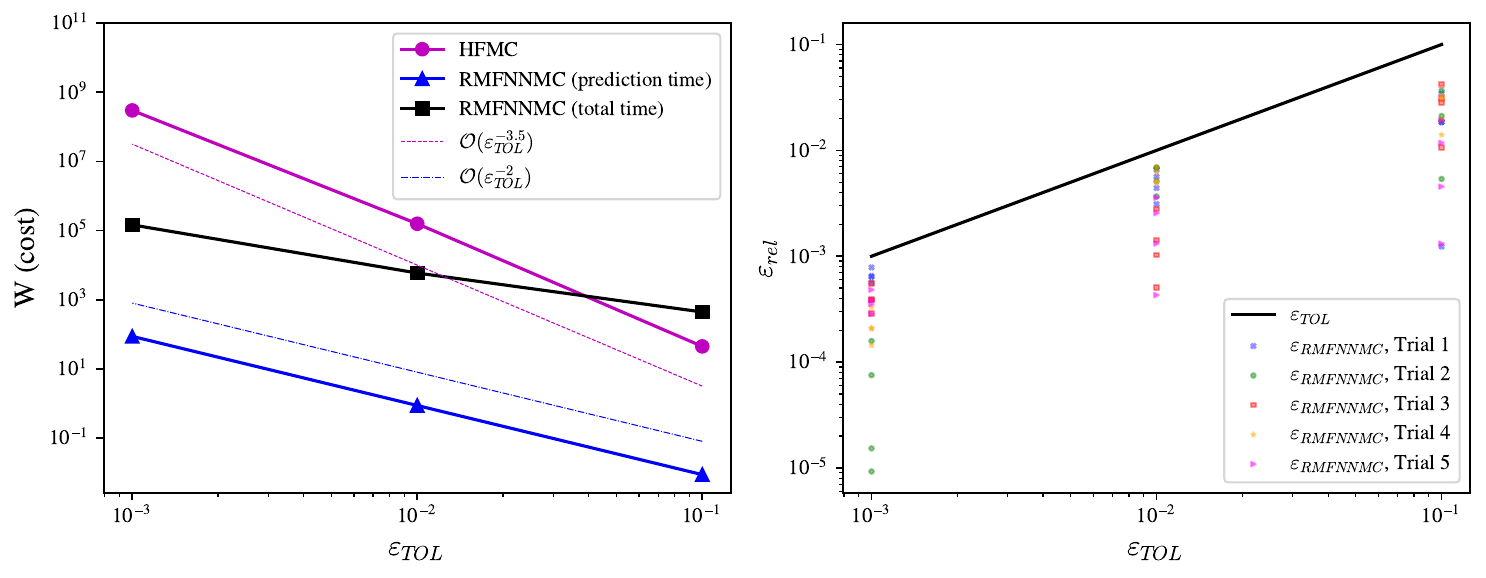}
\caption{CPU time and error versus tolerance. Left: for large tolerances the
  training cost is dominant, making the cost of
  RMFNNMC more than the cost of HFMC. However, as tolerance decreases,
  the cost of RMFNNMC approaches ${\mathcal
  O}(\varepsilon_{TOL}^{-2})$; Right: relative error over different training trials as a function of tolerance verifies that the
  tolerance is met with 1$\%$ failure probability; different markers indicate 4 Monte Carlo simulations from each training trial}
\label{Ex2_conv_tols}
\end{figure}

\bigskip
\noindent
{\bf MC sampling.} We now compute ${\mathbb E}[Q]$ by
\eqref{A_RMFNNMC} and \eqref{A_HFMC} and compare their complexities. 
Figure \ref{Ex2_conv_tols} (left) shows the CPU time as a function of
tolerance. The computational cost of a direct high-fidelity MC sampling is
proportional to $\varepsilon_{TOL}^{-3.5}$,
following \eqref{HFMC_cost_tol} and noting that the order of
accuracy of the finite difference scheme is $q=2$, and that the time-space dimension of the problem is $\gamma = 3$. 
On the other hand, if we only consider the evaluation cost of the
RMFNN algorithm constructed surrogate, excluding the training costs, the cost of the proposed RMFNN method is proportional to $\varepsilon_{TOL}^{-2}$ which
is much less than the cost of high-fidelity MC sampling. When adding the training
costs, we observe that although for large tolerances the training cost
is large, as the tolerance decreases the training costs become
negligible compared to the total CPU time. Overall, the cost of the
proposed method approaches ${\mathcal
  O}(\varepsilon_{TOL}^{-2})$ as tolerance
decreases, indicating orders of magnitude acceleration in computing
the expectation compared to high-fidelity MC sampling. 
This convergence rate can also be seen by \eqref{RMFNNMC_cost_tol},
noting that $\max(p + \gamma/q, 2+\hat{p}) = \max(0.2 + 1.5,
2+\hat{p}) = 2 + \hat{p}$, where $\hat{p}>0$ is very small. 
We note that while we observe significant cost reductions at the tested tolerances, training neural networks to achieve very small tolerances can be challenging, and what is required to do this in practice, be it additional training data or optimization time is not rigorously understood \cite{adcock2021gap}. Therefore, we do not assert that the demonstrated scaling will necessarily apply in these extremely small tolerance scenarios.
Finally, Figure \ref{Ex2_conv_tols} (right) shows 20 instances of the relative error as a function
of tolerance computed as 4 Monte Carlo simulations from each of the 5 independent network training runs. We see considerable variance in the network performance across different training runs, but even taking this variance into account, we meet the desired tolerance across all simulations.

\section{Conclusion}
\label{sec:conclusion}

In this work, we presented a residual multi-fidelity computational framework that leverages a pair of neural networks to efficiently construct surrogates for high-fidelity QoIs described by systems of ODEs/PDEs. 
Given a low-fidelity and a high-fidelity computational model, we first
formulate the relation between the two models in terms of a
possibly non-linear residual function that measures the discrepancy between the two model outputs. This modeling choice was motivated by a recently proved theoretical result \cite{Davis_Motamed:2024}, which guarantees the existence of ReLU networks approximating a large class of bounded target functions with minimal regularity assumptions to arbitrary tolerance with network complexity proportional to the uniform norm of the target function. This estimate provides general guidelines concerning when the developed modeling framework will be performative. 
When the residual function is small in uniform norm relative to the high-fidelity quantity, an approximating neural network $ResNN$ can be constructed with lower relative network complexity. We hypothesize that this network can be effectively trained using a sparse set of high- and low-fidelity data, aligning with the typical conditions of most multi-fidelity modeling problems. 
The trained network $ResNN$ is then used to efficiently generate additional high-fidelity data.
Finally, the entire set of available and newly generated high-fidelity data is used to train a deep network, $DNN$, which acts as a cost-efficient surrogate for the high-fidelity QoI.

We presented four numerical examples to demonstrate the power of the proposed framework. In Example~\ref{sec:num_ex_0}, we provided numerical justification for the inclusion of the low-fidelity model output in the input space of the residual function. We showed that this inclusion can yield a target function that is higher-dimensional, but with a simpler profile that can be learned by a network of lower complexity. In Example~\ref{sec:num_ex_1}, we conducted a bi-fidelity surrogate construction task using three approaches: 1) the RMFNN algorithm, 2) the MFNN method from \cite{MFNN_Motamed:20}, and 3) a pure high-fidelity neural network without leveraging lower-fidelity information. 
Our results demonstrated orders-of-magnitude reduction in generalization error for the RMFNN surrogate compared to the other two methods. This performance advantage was most pronounced with sparse high-fidelity data, supporting our hypothesis linking reduced network complexity to improved training on sparse data. Moreover, in a multi-fidelity context, the RMFNN framework makes a meaningful improvement over the residual learning framework as formulated in ResNet \cite{ResNets:2016}. Continuing, in examples \ref{sec:num_ex_2} and \ref{sec:num_ex_3} we used the RMFNN-constructed surrogate to approximate the expectation of our QoI via MC sampling. 
We demonstrated significant computational savings that increased as the desired error tolerance decreased, highlighting the efficiency of the RMFNN framework in reducing computational cost while maintaining accuracy.

One important future research direction concerns the extension of the RMFNN algorithm from bi-fidelity modeling to multi-fidelity modeling. The algorithm extends naturally in the case of an ensemble of lower-fidelity models that are strictly hierarchical in terms of their predictive accuracy for a given cost. In this case, we leverage a sequence of networks to learn a sequence of residual functions that efficiently generate additional high-fidelity data. The extension of the algorithm to more complex multilevel hierarchies of lower-fidelity models is an important open problem.

\section*{Declarations}

\begin{itemize}
\item Data availability: Data sets generated during the current study are available at \url{https://github.com/ondavis1997/residual_multifidelity} 
\item Author contribution: All authors contributed meaningfully to the present work. 
\end{itemize}

\bibliographystyle{unsrt}
\bibliography{rmfnn}

\end{document}